\documentclass[10pt,journal,compsoc]{IEEEtran}
\usepackage[latin9]{inputenc}
\usepackage{array}
\usepackage{float}
\usepackage{booktabs}
\usepackage{bm}
\usepackage{multirow}
\usepackage{amsmath}
\usepackage{graphicx}
\usepackage{setspace}
\usepackage{wasysym}
\usepackage{color}

\makeatletter

\floatstyle{ruled}
\newfloat{algorithm}{tbp}{loa}
\providecommand{\algorithmname}{Algorithm}
\floatname{algorithm}{\protect\algorithmname}

\ifCLASSOPTIONcompsoc
  \usepackage[nocompress]{cite}
\else
  \usepackage{cite}
\fi
\ifCLASSINFOpdf
\else
\fi
\usepackage{amsfonts}
\usepackage{nicefrac}
\usepackage{enumitem}
\usepackage{bookmark}

\usepackage{amsthm}

\newtheorem{innercustomgeneric}{\customgenericname}
\providecommand{\customgenericname}{}
\newcommand{\newcustomtheorem}[2]{%
  \newenvironment{#1}[1]
  {%
   \renewcommand\customgenericname{#2}%
   \renewcommand\theinnercustomgeneric{##1}%
   \innercustomgeneric
  }
  {\endinnercustomgeneric}
}

\newcustomtheorem{customthm}{Theorem}
\newcustomtheorem{customlemma}{Lemma}
\newcustomtheorem{customdef}{Definition}

\makeatother

\begin{document}
\title{Understanding the Disharmony between Weight Normalization Family and Weight Decay: $\epsilon-$shifted $L_2$ Regularizer}

\author{Xiang~Li, Shuo~Chen, Yan~Xia and Jian~Yang*
\IEEEcompsocitemizethanks{\IEEEcompsocthanksitem Xiang~Li, Shuo~Chen, and Jian~Yang are with the School of Computer Science and Engineering, Nanjing University of Science and Technology, Nanjing 210094, China (E-mail: (xiang.li.implus,~shuochen,~csjyang)@njust.edu.cn). Xiang~Li is also a visiting scholar at Momenta. Yan~Xia (E-mail:~xiayan@momenta.ai) is the research and development director of Momenta. (Corresponding~author: Jian~Yang)
}}\IEEEtitleabstractindextext{
\begin{abstract}
The merits of fast convergence and potentially better performance of the weight normalization family have drawn increasing attention in recent years. These methods use standardization or normalization that changes the weight $\boldsymbol{W}$ to $\boldsymbol{W}'$, which makes $\boldsymbol{W}'$ independent to the magnitude of $\boldsymbol{W}$. Surprisingly, $\boldsymbol{W}$ must be decayed during gradient descent, otherwise we will observe a severe under-fitting problem, which is very counter-intuitive since weight decay is widely known to prevent deep networks from over-fitting. Moreover, if we substitute (e.g., weight normalization) $\boldsymbol{W}'=\frac{\boldsymbol{W}}{||\boldsymbol{W}||}$ in the original loss function $\sum_{i}^{N}L(f( \boldsymbol{x}_i; \boldsymbol{W}'), y_i) + \frac{1}{2} \lambda ||\boldsymbol{W}'||^2$, it is observed that the regularization term $\frac{1}{2} \lambda ||\boldsymbol{W}'||^2$ will be canceled as a constant $\frac{1}{2} \lambda$ in the optimization objective. Therefore, to decay ${\boldsymbol{W}}$, we need to explicitly append this term: $\frac{1}{2} \lambda ||{\boldsymbol{W}}||^2$. In this paper, we \emph{theoretically} prove that $\frac{1}{2} \lambda ||{\boldsymbol{W}}||^2$ merely modulates the effective learning rate for improving objective optimization, and  has no influence on generalization when the weight normalization family is compositely employed. Furthermore, we also expose several critical problems when introducing weight decay term to weight normalization family, including the missing of global minimum and training instability. To address these problems, we propose an $\epsilon-$shifted $L_2$ regularizer, which shifts the $L_2$ objective by a positive constant $\epsilon$. Such a simple operation can theoretically guarantee the existence of global minimum, while preventing the network weights from being too small and thus avoiding gradient float overflow. It significantly improves the training stability and can achieve slightly better performance in our practice. The effectiveness of $\epsilon-$shifted $L_2$ regularizer is comprehensively validated on the ImageNet, CIFAR-100, and COCO datasets. Our codes and pretrained models will be released in https://github.com/implus/PytorchInsight.
\end{abstract}


}\maketitle

\section{Introduction}

\IEEEPARstart{T}{he} normalization methodologies on \emph{features} have made great progress in recent years, with the introduction of BN \cite{ioffe2015batch}, IN \cite{ulyanov2016instance}, LN \cite{ba2016layer}, GN \cite{wu2018group} and SN \cite{luo2018differentiable}. These methods mainly focus on a zero mean and unit variance normalization operation on a specific dimension (or multiple dimensions) of \emph{features}, which makes deep neural architectures \cite{he2016deep,he2016identity,huang2017densely,wang2018mixed} much easier to optimize, leading to robust solutions with favorable generalization performance.

Beyond \emph{feature} normalization, there is an increasing interest on the normalization of network \emph{weights}. Weight Normalization (WN) \cite{salimans2016weight} first separates the learning of the length and direction of weights, and it performs satisfactorily on several relatively small datasets. In some contexts of generative adversarial networks (GAN) \cite{goodfellow2014generative}, Weight Normalization with Translated ReLU \cite{xiang2017effects} is shown to achieve superior results. Later, Centered Weight Normalization (CWN) \cite{huang2017centered} further powers WN by additionally centering their input weights, ulteriorly improving the conditioning and convergence speed. Recently, very similar to CWN, Weight Standardization (WS) \cite{weightstandardization} aims to standardize the weights with zero mean and unit variance. On the large-scale tasks (ImageNet \cite{deng2009imagenet} classification/COCO \cite{lin2014microsoft} detection), WS further enhances optimization convergence and generalization performance, under the cooperation of feature normalizations such as GN and BN.

\begin{figure}[t]
	\begin{center}
		\setlength{\fboxrule}{0pt}
		\fbox{\includegraphics[width=0.44\textwidth,height=0.28\textwidth]{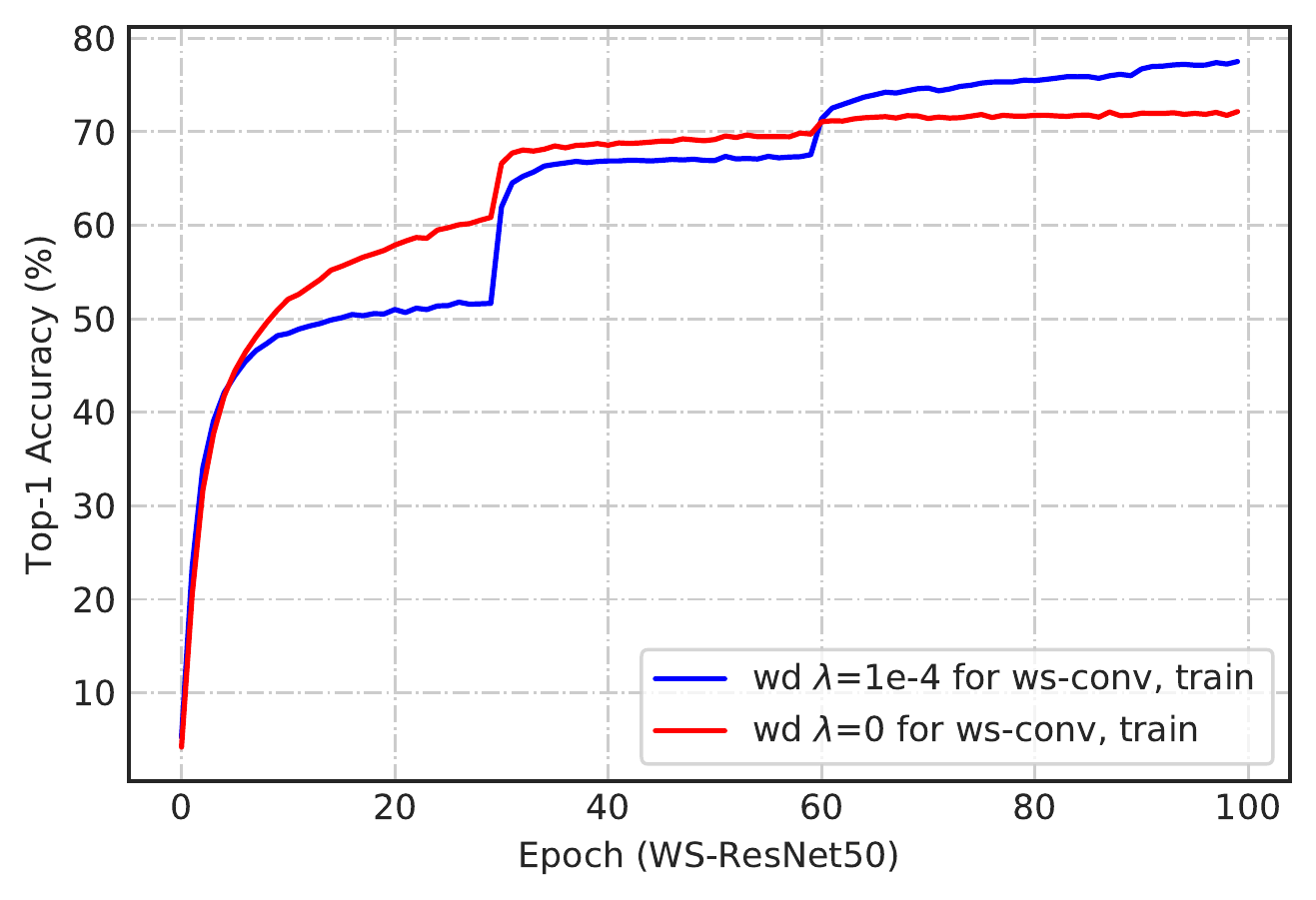}}
	\end{center}	
	\vspace{-16pt}
	\caption{Counter-intuitive performance degradation (under-fitting) by setting weight decay parameter $\lambda$ to 0 for WS-equipped convolutions. Top-1 Accuracy via single 224$\times$ crop on the ImageNet training set is plotted.}
	\label{fig_wd}
\end{figure}

In terms of weight normalization family, despite its appealing success, there is still one confusing mystery -- the disharmony between weight normalization family and weight decay \cite{krogh1992simple}. Note that weight decay is widely interpreted as a form of $L_2$ regularization \cite{loshchilov2017fixing} because it can be derived from the gradient of the $L_2$ norm of the weights \cite{loshchilov2018decoupled}. Specifically, we consider training a single-layer single-output neural network $f(\boldsymbol{x}; {\boldsymbol{W}'}),$ where $\boldsymbol{x},{\boldsymbol{W}'} \in\mathcal{R}^{n} $ with the following loss function to be minimized:
\begin{equation}
\hat{\mathcal{L}}({\boldsymbol{W}}') = \sum_{i}^N L(f(\boldsymbol{x}_i; {\boldsymbol{W}}'), y_i) + \frac{1}{2} \lambda ||{\boldsymbol{W}}'||^2.
\label{eq1}
\end{equation}
In Eq.~(\ref{eq1}), $N$ denotes the number of training samples, $\hat{\mathcal{L}}$ consists of task-related loss $L(\cdot,\cdot)$ w.r.t. the input/label pair $(\boldsymbol{x}_i, y_i)$, and the regularization term $\frac{1}{2} \lambda ||{\boldsymbol{W}}'||^2$ with a constant $\lambda$ to balance against $L(\cdot,\cdot)$. For simplicity, we use weight normalization to re-parameterize ${\boldsymbol{W}}'$, regardless the learning of its length. By substituting ${\boldsymbol{W}}' = \frac{{\boldsymbol{W}}}{||{\boldsymbol{W}}||}$ in Eq. (\ref{eq1}), we can get $\sum_{i}^N L(f(\boldsymbol{x}_i; \frac{{\boldsymbol{W}}}{||{\boldsymbol{W}}||}), y_i) + \frac{1}{2} \lambda $, which is equivalent to minimize the following function
\begin{equation}
\sum_{i}^N L(f(\boldsymbol{x}_i; \frac{{\boldsymbol{W}}}{||{\boldsymbol{W}}||}), y_i).
\label{eq2}
\end{equation}
Interestingly, the weight decay term has indeed \emph{disappeared}. In the case of WS, we can get similar conclusions by replacing 
${\boldsymbol{W}}' = \frac{{\boldsymbol{W}} -\overline{{ W}}}{\sqrt{\frac{||{\boldsymbol{W}} - \overline{{ W}}||^2 }{n} }}$:
\begin{equation}
\sum_{i}^N L(f(\boldsymbol{x}_i; \frac{{\boldsymbol{W}} -\overline{{ W}}}{\sqrt{\frac{||{\boldsymbol{W}} - \overline{{ W}}||^2 }{n} }}), y_i),
\label{eq3}
\end{equation}
where $\overline{{ W}} = \frac{\sum_i^n { W}_i}{n}$. It probably makes sense since weight decay will not take effect on a fixed distribution of normalized weights. However, when we apply Eq. (\ref{eq3}) to WS-equipped ResNet-50 (WS-ResNet-50) on ImageNet dataset, i.e., setting weight decay ratio $\lambda$ to 0 for all WS-equipped convolutions, we observe a severe degradation with significant performance drop in training set (Fig. \ref{fig_wd}). It is incredibly \emph{strange} that weight decay is known to prevent the training from over-fitting the data, but it appears that, removing the weight decay instead puts the network into a serious under-fitting, which is very counter-intuitive.

To answer above questions, in this paper, we first prove that in Eq. (2) (and Eq. (3)), the addition of weight decay term of ${\boldsymbol{W}}$ does not change the optimization goal. Therefore, weight decay loses its original role that finds a better generalized solution by traditionally introducing a different loss part against the task-related one. At the same time, basing on the derivation of the gradient formula of ${\boldsymbol{W}}$, we further prove that weight decay only takes effect in modulating the effective learning rate to help the gradient descent process when the weight normalization family is employed simultaneously, and empirically demonstrate how it adjusts the effective learning rate. Interestingly, we also get an additional empirical discovery: training a network with weight normalization and weight decay implicitly includes an approximate warmup \cite{goyal2017accurate} process, which probably explains the slightly improved performance on the original baselines.

The current common and default operation \cite{salimans2016weight,weightstandardization} to optimize networks with weight normalization family is to continue to preserve the traditional decay term of ${\boldsymbol{W}}$ for better convergence that comes from ensuring the stable effective learning rate, i.e., to explicitly add $\frac{1}{2} \lambda  ||{\boldsymbol{W}}||^2$ on Eq. (\ref{eq2}):
\begin{equation}
\sum_{i}^N L(f(\boldsymbol{x}_i; \frac{{\boldsymbol{W}}}{||{\boldsymbol{W}}||}), y_i) + {\frac{1}{2} \lambda ||{\boldsymbol{W}}||^2}.
\label{eq4}
\end{equation}
However, there are many potential problems in taking the final optimization objective as Eq. (\ref{eq4}). First, we prove that Eq. (\ref{eq4}) has no global minimum theoretically. In addition, the improper selection of $\lambda$ will push the magnitude ($||\boldsymbol{W}||$) of weights to 0 and easily lead to training failures due to gradient float overflow (as the corresponding gradient is propotional to $\frac{1}{||\boldsymbol{W}||}$), especially for certain adaptive gradient methods (e.g., Adam \cite{kingma2014adam}) which accumulates the square of gradients. 

To address these problems, we propose a very simple yet effective $\epsilon-$shifted $L_2$ regularizer, which shifts the $L_2$ objective by a positive constant $\epsilon$. The shifted $\epsilon$ prevents network weights from being too small, thus it will directly avoid gradient float overflow risks. Such a simple operation can theoretically guarantee the existence of global minimum, whilst greatly improving the training stability. Beyond the training stability, it further brings gains on performance over a wide range of architectures, probably due to its dynamic decay machanism, which we will discuss later. The effectiveness of our method is comprehensively demonstrated by experiments on the ImageNet \cite{deng2009imagenet}, CIFAR-100 \cite{krizhevsky2009learning} and COCO \cite{lin2014microsoft} datasets.

To summarize our contributions:
\begin{itemize}
	\item We thoroughly analyze the disharmony between weight normalization family and weight decay, and expose the critical problems including lack of global minimum and training instability, which are caused by the optimization of weight decay term in the final loss objective when weight normalization is simultaneously applied. 
	
	\item We theoretically prove that weight decay loses the ability to enhance generalization in the weight normalization family, and only plays a role in regulating effective learning rate to help training. We demonstrate that when optimizing with SGD, the weight decay term can be cancelled by simply scaling the learning rate with a constant at each gradient descent step, where the constant is only determined by the hyper-parameters and irrelevant to the training process.
	
	\item We propose a simple yet effective $\epsilon-$shifted $L_2$ regularizer to overcome the problems via introducing weight decay into weight normalization family, which significantly improves the training stability whilst achieving better performance over a large range of network architectures on both classification and detection tasks.
\end{itemize}

\section{Related Works}
\noindent \textbf{Weight Normalization Family:} Weight Normalization (WN) \cite{salimans2016weight} takes the first attempt to reparameterize weights by the separation of direction $\frac{{\boldsymbol{W}}}{||{\boldsymbol{W}}||}$ and length $g$:
\begin{equation}
{\boldsymbol{W}}' = g \frac{{\boldsymbol{W}}}{||{\boldsymbol{W}}||}.
\end{equation}
The normalization operation participates in the gradient flow, resulting in accelerated convergence of stochastic gradient descent optimization. WN shows certain advantages in some tasks of supervised image recognition, generative modelling, and deep reinforcement learning. However, \cite{gitman2017comparison} points out that in the large-scale ImageNet dataset, the final test accuracy of WN is significantly lower ($\sim$ 6\%) than that of BN \cite{ioffe2015batch}. Later, Centered Weight Normalization (CWN) is proposed to further improve the conditioning and accelerate the convergence of training deep networks. The central idea of CWN is an additional centering operation based on WN:
\begin{equation}
{\boldsymbol{W}}' = g \frac{{\boldsymbol{W}} - \overline{{{W}}}}{|| {\boldsymbol{W}} - \overline{{{W}}}||}.
\end{equation}
Recently, in order to alleviate the problem of degraded performance of GN \cite{wu2018group}, Weight Standardization (WS) \cite{weightstandardization} is proposed, which is very close to CWN but with the learning length $g$ removed:
\begin{equation}
{\boldsymbol{W}}' = \frac{{\boldsymbol{W}} -\overline{{{W}}}}{\sqrt{\frac{||{\boldsymbol{W}} - \overline{{{W}}}||^2 }{n} }},
\end{equation}
WS is recommended to cooperate with feature normalization methods (such as GN and BN), which leads to further enhanced performance in large-scale tasks and can significantly accelerate the convergence. Introducing WS on the basis of GN or BN can consistently bring gains to multiple downstream visual tasks, including image classification, object detection, instance segmentation, video recognition, semantic segmentation, and point cloud recognition. In this paper, we mainly focus on the weight normalization family and conduct a series of analyses on their properties, especially on their relations to weight decay.

\noindent \textbf{Weight Decay:} weight decay can be traced back to \cite{krogh1992simple}, which is defined as multiplying each weight in the gradient descent at each epoch by a factor $\lambda (0 < \lambda < 1)$. In the Stochastic Gradient Descent (SGD) setting, weight decay is widely interpreted as a form of $L_2$ regularization \cite{ng2004feature} because it can be derived from the gradient of the $L_2$ norm of the weights \cite{loshchilov2018decoupled}. It is known to be beneficial for the generalization of neural networks. Recently, \cite{zhang2018three} identify three distinct mechanisms by which weight decay improves generalization: increasing the effective learning rate for BN, reducing the Jacobian norm, and reducing the effective damping parameter. Similarly, a series of recent work \cite{van2017l2,hoffer2018norm} also demonstrates that when using BN, weight decay improves optimization only by fixing the norm to a small range of values, leading to a more stable step size for the weight direction. Although related, these works differ from our work in at least four aspects: 1) they mainly focus on the discussion between the feature normalization (especially BN) and weight decay, whilst we are the first to give a thorough analysis on the disharmony between weight normalization family and weight decay; 2) they solely demonstrate \emph{empirical} results that the accuracy gained by using weight decay can be achieved without it, but only by adjusting the learning rate. However, we give \emph{theoretical} proof and derive how to linearly scale the learning rate at each step, which is also purely determined by the training hyper-parameters; 3) they fail to discover the problems by introducing  weight decay into the loss objective with normalized weights, which is heavily revealed and discussed in this article; 4) although weight decay has several potential problems with normalization methods, they have not proposed a solution to replace weight decay. In contrast, our proposed $\epsilon$-shifted $L_2$ regularizer can successfully guarantee the global minimum and training stability to overcome the existing drawbacks, whilst achieving superior performance over a range of tasks.

\begin{figure*}[t]
	\begin{center}
		\setlength{\fboxrule}{0pt}
		\fbox{\includegraphics[width=0.82\textwidth,height=0.28\textwidth]{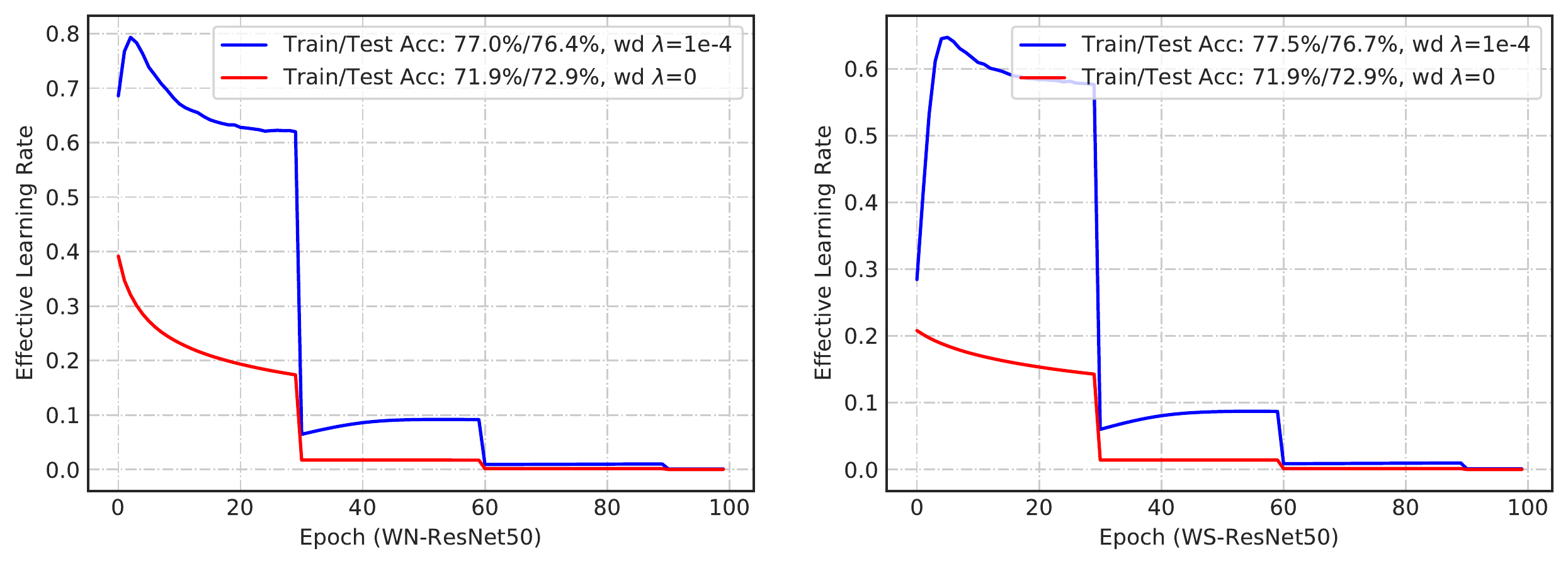}}
	\end{center}	
	\vspace{-16pt}
	\caption{The comparisons of effective learning rate for different weight decay $\lambda$ at each epoch during the optimization of WN-ResNet (left) and WS-ResNet (right) on the ImageNet dataset, where the effective learning rates of their first convolutional layer are depicted. The {\textcolor{blue}{blue}} curve ensures a larger effective learning rate which helps the networks converge.}
	\label{fig_elr}
\end{figure*}

\section{Roles of Weight Decay in Weight Normalization Family}
In this section, we explain the \emph{roles} of weight decay in weight normalization family in details. The theoretical analyses on the roles of weight decay help to understand why weight decay loses the ability to enhance the generalization, but controls the effective learning rate to help the training of deep networks.
\subsection{Weight Decay Doesnot Change Optimization Goal}
We first prove that in the networks equipped with weight normalization family, the introduction of weight decay does not change the goal of optimization, indicating that weight decay faithfully brings  no additional generalization benefits. For analyses, we simply use the concepts of variable decomposition. Specifically, we choose two representative methods from weight normalization family, namely WN and WS, and discuss each in turn. Note that for simplicity, we ignore the learning of the length $g$ in WN in the following analyses and experiments.

In WN, we aim to prove that {minimizing}
\begin{equation}
\hat{\mathcal{L}}({\boldsymbol{W}}) = \sum_{i}^N L(f(\boldsymbol{x}_i; \frac{{\boldsymbol{W}}}{||{\boldsymbol{W}}||}), y_i) + \frac{1}{2} \lambda ||{\boldsymbol{W}}||^2,
\label{eq5}
\end{equation}
is equal to {minimizing}
\begin{equation}
\mathcal{L}({\boldsymbol{W}}) = \sum_{i}^N L(f(\boldsymbol{x}_i; \frac{{\boldsymbol{W}}}{||{\boldsymbol{W}}||}), y_i).
\label{eq6}
\end{equation}
{Specifically,} we let ${\boldsymbol{A}} = \frac{{\boldsymbol{W}}}{||{\boldsymbol{W}}||}$ and $k = ||{\boldsymbol{W}}|| (k > 0)$, and decompose the direction and length of ${\boldsymbol{W}}$ as two independent variables. Then the {objectives} of Eq. (\ref{eq5}) and (\ref{eq6}) can be rewritten as 
\begin{align}
& \min_{{\boldsymbol{A}},k}\hat{\mathcal{L}}({\boldsymbol{A}},k)=\sum_{i}^N L(f(\boldsymbol{x}_{i};{\boldsymbol{A}}),y_{i})+\frac{1}{2}\lambda k^{2},\nonumber \\
& s.t.\thinspace\thinspace\thinspace||{\boldsymbol{A}}||=1,k>0,  \label{eq7}
\end{align}
and
\begin{align}
&
\min_{{\boldsymbol{A}}}\mathcal{L}({\boldsymbol{A}}) = \sum_{i}^N L(f(\boldsymbol{x}_i; {\boldsymbol{A}}), y_i), \nonumber\\
& s.t.\ \  ||{\boldsymbol{A}}||=1, \label{eq8}
\end{align}
respectively. Since ${\boldsymbol{A}}$ and $k$ are two independent variables, we have 
\begin{align}
\min_{{\boldsymbol{A}},k}\hat{\mathcal{L}}({\boldsymbol{A}},k) & =\min_{{\boldsymbol{A}}}\sum_{i}^N L(f(\boldsymbol{x}_{i};{\boldsymbol{A}}),y_{i})+\min_{k}\frac{1}{2}\lambda k^{2}\nonumber \\
& =\min_{{\boldsymbol{A}}}{\mathcal{L}}({\boldsymbol{A}})+\min_{k}\frac{1}{2}\lambda k^{2},\label{eq9}
\end{align}
which shows that minimizing $\hat{\mathcal{L}}$ actually contains the task of minimizing ${\mathcal{L}}$, and it completes the proof. Similarly, in WS we can further decompose the mean and variance of ${\boldsymbol{W}}$ by letting ${\boldsymbol{B}} = \frac{{\boldsymbol{W}} -\overline{{ W}}}{\sqrt{\frac{||{\boldsymbol{W}} - \overline{{ W}}||^2 }{n} }}, m = \overline{{ W}}$ and $v = \sqrt{\frac{||{\boldsymbol{W}} - \overline{{ W}}||^2 }{n}} (v > 0)$, where ${\boldsymbol{B}}, m, v$ are also mutually independent. Again we can have 
\begin{equation}
\min_{{\boldsymbol{B}}, m, v}\hat{\mathcal{L}}({\boldsymbol{B}}, m, v) = \min_{{\boldsymbol{B}}}{\mathcal{L}}({\boldsymbol{B}}) + \min_{m, v} \frac{1}{2} \lambda n(m^2 + v^2). \label{eq10}\\
\end{equation}

Therefore, according to the above analyses, for the networks with weight normalization family employed, the introduction of weight decay does not essentially change the {learning objective}, which {implies} that it {takes no effect on the network generalization capability}.

\subsection{Weight Decay Ensures Effective Learning Rate\label{sec2.2}}
Since weight decay does not bring a regularization effect to a network with weight normalization family, why is it indispensable in the training process? The central reason is that weight decay helps to control the effective learning rate in a stable and reasonable range. Taking WN as an example, we can derive the gradient of ${\boldsymbol{W}}$ as (the deviate to Eq. (\ref{eq6})):
\begin{equation}
\frac{\partial \mathcal{L}}{\partial {\boldsymbol{W}}} = \frac{\partial \mathcal{L}}{\partial {\boldsymbol{A}}} * (\frac{\bf{1}}{||{\boldsymbol{W}}||} - \frac{{\boldsymbol{W}} * {\boldsymbol{W}}}{||{\boldsymbol{W}}||^3}),
\label{eqgradient}
\end{equation}
where $*$ denotes the element-wise product. If we consider one gradient descent update at $t$ step for an element in ${\boldsymbol{W}}$, i.e., ${ W}_i$, with the use of learning rate $\eta$, we have:
\begin{equation}
{ W}_i^{t+1} =  { W}_i^{t} - \frac{\eta}{||{\boldsymbol{W}}^t||}(1 - \frac{({ W}_i^{t})^2}{\sum_{j}^n({ W}_j^{t})^2})\frac{\partial \mathcal{L}}{\partial { A}_i}. \label{e}
\end{equation}

Eq.~(\ref{e}) demonstrates that even if ${\boldsymbol{A}} = \frac{{\boldsymbol{W}}}{||{\boldsymbol{W}}||}$ is fixed, the gradient update in terms of ${\boldsymbol{W}}$ can vary according to its magnitude $||{\boldsymbol{W}}||$. The reason is that fixed ${\boldsymbol{A}}$ can only leads to fixed $\frac{\partial \mathcal{L}}{\partial {{A}}_i}$ and $(1 - \frac{({{W}}_i^{t})^2}{\sum_{j}^n({{W}}_j^{t})^2})$ term. Consequently, the entire update step size can be determined by $\frac{\eta}{||{\boldsymbol{W}}||}$, which exactly controls the effective learning rate similarly defined as in \cite{van2017l2,hoffer2018norm}. Here we have two conclusions:

\begin{itemize}
	\item If we do not limit $||{\boldsymbol{W}}||$ during the update process, the weights can grow unbounded $||{\boldsymbol{W}}|| \to \infty$, and the effective learning rate goes to 0 ($\frac{\eta}{||{\boldsymbol{W}}||} \to 0$).
	
	\item On the contrary, if we decay $||{\boldsymbol{W}}||$ too much during the optimization ($||{\boldsymbol{W}}|| \to 0$), the effective learning rate will grow unbounded ($\frac{\eta}{||{\boldsymbol{W}}||} \to \infty$), which leads to gradient float overflow and training failures. This is part of the motivation of our proposed method and we will discuss it in details later.
\end{itemize}

The similar analysis can be conducted in the case of WS, where one update step is:
\begin{align}
 & {W}_{i}^{t+1}\nonumber \\
 & ={W}_{i}^{t}\!-\!\frac{\eta}{\sqrt{\frac{||{\boldsymbol{W}}^{t}\!-\!\overline{{W}^{t}}||^{2}}{n}}}(1\!-\!\frac{1}{n}\!-\!\frac{({W}_{i}^{t}\!-\!\overline{{W}^{t}})^{2}}{\sum_{j}^{n}({W}_{j}^{t}\!-\!\overline{{W}^{t}})^{2}})\frac{\partial\mathcal{L}}{\partial{B}_{i}}, \label{eq16}
\end{align}
which has $\frac{\eta}{\sqrt{\frac{||{\boldsymbol{W}} - \overline{{{W}}}||^2 }{n}}}$ term as its effective learning rate.

To prove that weight decay only ensures effective learning rate, we conduct theoretical analyses as follows:

\begin{customthm}{1} Using SGD, the training trajectory of $\hat{\mathcal{L}}$ (Eq. (\ref{eq5}) with weight decay) can be completely reproduced by simply scaling the learning rate at each step when optimizing $\mathcal{L}$ (Eq. (\ref{eq6}) without weight decay).
\end{customthm}

\begin{figure}[t]
	\begin{center}
		\setlength{\fboxrule}{0pt}
		\fbox{\includegraphics[width=0.44\textwidth]{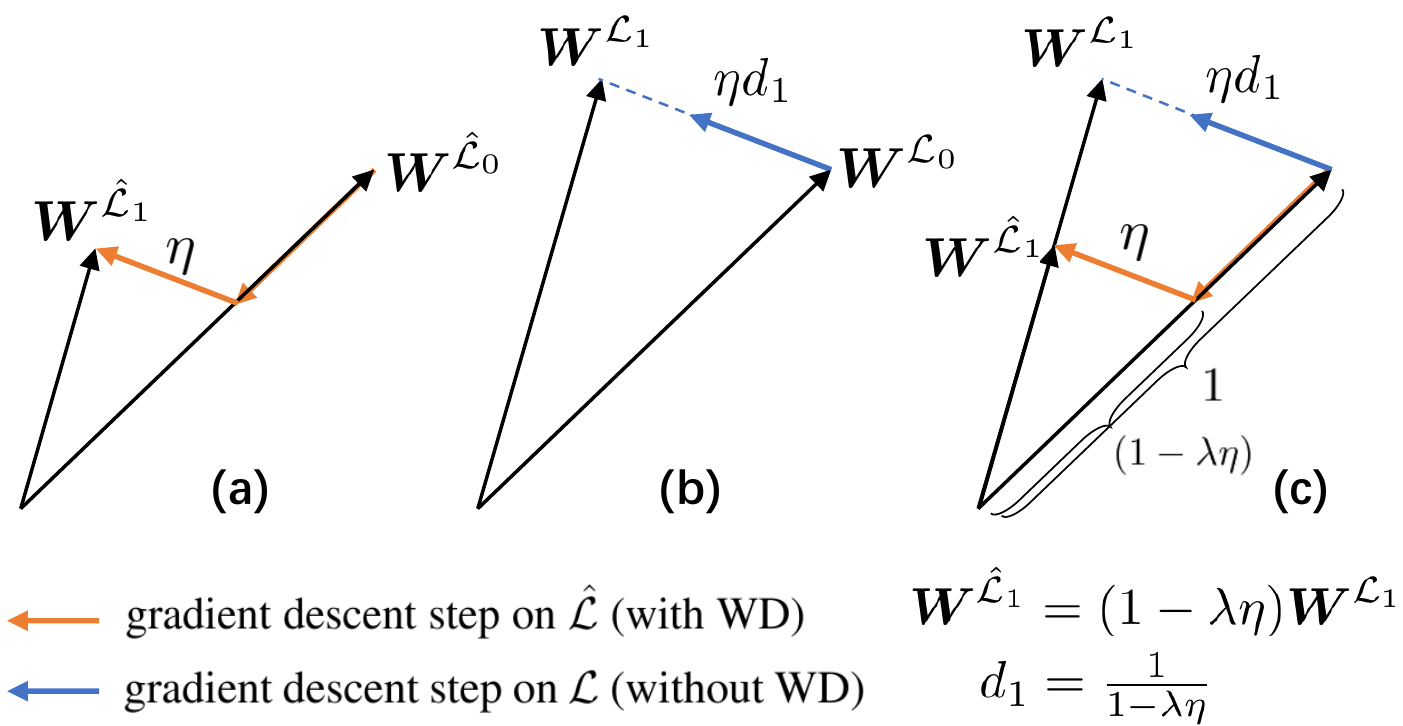}}
	\end{center}	
	\vspace{-12pt}
	\caption{Illustration of proving that weight decay can be entirely replaced by modulating the learning rate (under the setting of $p_0 = 1$ in this example). \textbf{(a)} denotes one update with weight decay. \textbf{(b)} denotes one update without weight decay. \textbf{(c)} shows the similar triangle relationship between these two updates, where the equations ${\boldsymbol{W}}^{\hat{\mathcal{L}}_{1}} = (1 - \lambda\eta){\boldsymbol{W}}^{{\mathcal{L}}_{1}}$ and $d_1 = \frac{1}{1 - \lambda\eta}$ can be easily derived.}
	\label{fig_gd}
\end{figure}

\begin{proof} Here we focus on the normalized variables to represent the training trajectory, e.g., ${\boldsymbol{A}}$ (or ${\boldsymbol{B}}$), as they are the ultimate weights with which networks use to operate. In the case of WN, we suppose optimizing $\hat{\mathcal{L}}({\boldsymbol{A}}, k)$ and ${\mathcal{L}}(A)$ take $T$ steps in total, and at each step, we feed the same data batch to both of them. For the ease of reference, the corresponding variables at $t$ step for optimizing $\hat{\mathcal{L}}$ are marked with superscript $\hat{\mathcal{L}}_t$, e.g., ${\boldsymbol{W}}^{\hat{\mathcal{L}}_t}$. Such notations are kept similarly in optimizing ${\mathcal{L}}$, e.g., ${\boldsymbol{W}}^{{\mathcal{L}}_t}$. We further assume that two optimization processes start from the same initial weights ${\boldsymbol{A}}^{\hat{\mathcal{L}}_0} = {\boldsymbol{A}}^{{\mathcal{L}}_0}$, which also means ${\boldsymbol{W}}^{\hat{\mathcal{L}}_0} = p_0 {\boldsymbol{W}}^{{\mathcal{L}}_0}$, where $p_t$ is a scale between ${\boldsymbol{W}}^{\hat{\mathcal{L}}_t}$ and ${\boldsymbol{W}}^{{\mathcal{L}}_t}$ (if $p_t$ exists). Specifically, we aim to prove that there exists a sequence of $\{d_1, ..., d_T\}$ as multipliers (note that $d$ must be independent of the training process) for learning rate $\eta$ during optimizing $\mathcal{L}$, and it ensures ${\boldsymbol{A}}^{\hat{\mathcal{L}}_t} = {\boldsymbol{A}}^{{\mathcal{L}}_t}$ for every step $t$ from $\{1, ..., T\}$. We take the standard SGD \cite{bottou2010large,ruder2016overview} for analysis via mathematical induction:

1) As stated in the assumption, we have ${\boldsymbol{A}}^{\hat{\mathcal{L}}_0} = {\boldsymbol{A}}^{{\mathcal{L}}_0}$ hold for $t = 0$, indicating ${\boldsymbol{W}}^{\hat{\mathcal{L}}_0} = p_{0} {\boldsymbol{W}}^{{\mathcal{L}}_0}$. Here we do not have $d_0$ since  the gradient descent step does not start in the initialization phase.

2) Suppose ${\boldsymbol{A}}^{\hat{\mathcal{L}}_q} = {\boldsymbol{A}}^{{\mathcal{L}}_q}$ holds for $t = q$ with ${\boldsymbol{W}}^{\hat{\mathcal{L}}_q} = p_{q} {\boldsymbol{W}}^{{\mathcal{L}}_q}$, we needs to prove ${\boldsymbol{A}}^{\hat{\mathcal{L}}_{q+1}} = {\boldsymbol{A}}^{{\mathcal{L}}_{q+1}}$ under certain expression of $d_q$. Let us expand ${\boldsymbol{W}}^{\hat{\mathcal{L}}_{q+1}}$ and ${\boldsymbol{W}}^{{\mathcal{L}}_{q+1}}$ by performing one gradient descent step:
\begin{align}
& {\boldsymbol{W}}^{\hat{\mathcal{L}}_{q+1}}\nonumber \\
& =\!(1\!-\!\lambda\eta){\boldsymbol{W}}^{\hat{\mathcal{L}}_{q}}\!-\!\frac{\eta}{||{\boldsymbol{W}}^{\hat{\mathcal{L}}_{q}}||}({\bf 1}\!-\!\frac{{\boldsymbol{W}}^{\hat{\mathcal{L}}_{q}}\!*\!{\boldsymbol{W}}^{\hat{\mathcal{L}}_{q}}}{||{\boldsymbol{W}}^{\hat{\mathcal{L}}_{q}}||^{2}})*\frac{\partial\mathcal{L}}{\partial{\boldsymbol{A}}^{\hat{\mathcal{L}}_{q}}}\nonumber \\
& =(1\!\!-\!\!\lambda\eta)p_{q}{\boldsymbol{W}}^{{\mathcal{L}}_{q}}\!-\!\frac{\eta}{p_{q}||{\boldsymbol{W}}^{{\mathcal{L}}_{q}}||}({\bf 1}\!\!-\!\!\frac{{\boldsymbol{W}}^{{\mathcal{L}}_{q}}\!*\!{\boldsymbol{W}}^{{\mathcal{L}}_{q}}}{||{\boldsymbol{W}}^{{\mathcal{L}}_{q}}||^{2}})*\frac{\partial\mathcal{L}}{\partial{\boldsymbol{A}}^{{\mathcal{L}}_{q}}}, \label{eq14}
\end{align}
and
\begin{equation}
{\boldsymbol{W}}^{{\mathcal{L}}_{q+1}}= {\boldsymbol{W}}^{{\mathcal{L}}_{q}} - \frac{\eta d_q}{||{\boldsymbol{W}}^{{\mathcal{L}}_{q}}||}({\bf 1} - \frac{{\boldsymbol{W}}^{{\mathcal{L}}_{q}} * {\boldsymbol{W}}^{{\mathcal{L}}_{q}}}{||{\boldsymbol{W}}^{{\mathcal{L}}_{q}}||^2})*\frac{\partial \mathcal{L}}{\partial {\boldsymbol{A}}^{{\mathcal{L}}_{q}}}.
\label{eq15}
\end{equation}
Therefore, it is very obvious to deduce from Eq. (\ref{eq14}) and (\ref{eq15}) that when $d_q = \frac{1}{p_q^2 (1 - \lambda \eta)}$, we can have 
\begin{equation}
{\boldsymbol{W}}^{\hat{\mathcal{L}}_{q+1}} = (1 - \lambda\eta)p_q{\boldsymbol{W}}^{{\mathcal{L}}_{q+1}},
\end{equation}
which consequently leads to ${\boldsymbol{A}}^{\hat{\mathcal{L}}_{q+1}} = {\boldsymbol{A}}^{{\mathcal{L}}_{q+1}}$ and thereby it completes the proof. At the same time, we can also derive the recursive formula for $p$:
\begin{equation}
p_{q+1} = (1 - \lambda \eta)p_{q}.
\label{eq17}
\end{equation}
Given the sequence of $p$ generated from Eq. (\ref{eq17}), the resulted sequence $\{d_1, ..., d_T\}$ of $d$ then becomes $\{\frac{1}{p_1^2 (1 - \lambda \eta)}, ..., \frac{1}{p_T^2 (1 - \lambda \eta)}\}$ finally. The similar deductions can be carried out for the case of WS. To give a better illustration of the proof, we let $p_0 = 1$ and demonstrate the first gradient descent update of $\hat{\mathcal{L}}$ and ${\mathcal{L}}$ in Fig. \ref{fig_gd}. It is easy to see that ${\boldsymbol{W}}^{\hat{\mathcal{L}}_{1}}$ and ${\boldsymbol{W}}^{{\mathcal{L}}_{1}}$ form a similar triangle relationship and their scale factor is only determined by the hyper-parameter $\lambda$ and $\eta$.
\end{proof}

\begin{figure}[t]
	\begin{center}
		\setlength{\fboxrule}{0pt}
		\fbox{\includegraphics[width=0.44\textwidth,height=0.28\textwidth]{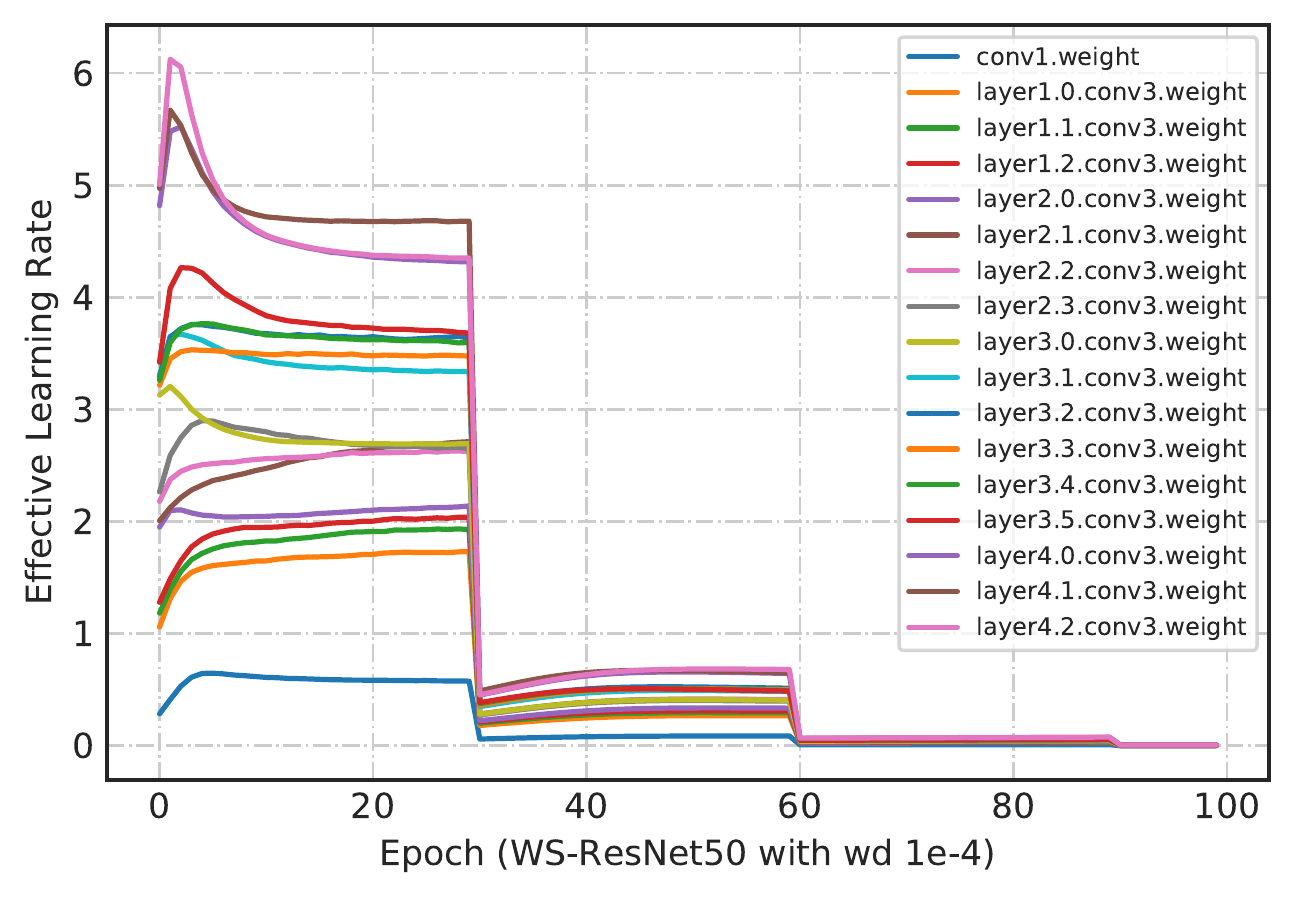}}
	\end{center}	
	\vspace{-16pt}
	\caption{The curves of mean effective learning rate of multiple WS-equipped convolutional layers. We observe an interesting warmup phenomenon in the initial training stage. In the form of ``layer$a$.$b$.conv$c$'', ``$a$'' denotes the stage number, ``$b$'' denotes the bottleneck number, ``$c$'' represents the order of convolutional layer inside this bottleneck. For reference, the first ``conv1'' means the first convolutional layer, which is exactly the same with the blue curve in WS-ResNet-50 of Fig. \ref{fig_elr}.}
	\label{fig_elrall}
\end{figure}

According to the above analyses, we conclude that for networks with weight normalization family, weight decay only takes effect in modulating effective learning rate, and theoretically, we can replace it simply by adjusting the learning rate in each iteration with a calculated ratio which is only related to the hyper-parameter $\lambda$ and $\eta$. 

To show how weight decay regulates the effective learning rate, we plot the mean effective learning rate of all the filters from the first layer of WN-ResNet-50 and WS-ResNet-50 throughout the training process in Fig. \ref{fig_elr}, where 0 and 1e-4 weight decay ratios are applied to the corresponding convolutional layers respectively. It is observed that the effective learning rate is appropriately controlled in a relatively large range with weight decay applied, which leads to a better optimized solution. 

\begin{figure}[t]
	\begin{center}
		\setlength{\fboxrule}{0pt}
		\fbox{\includegraphics[width=0.44\textwidth,height=0.28\textwidth]{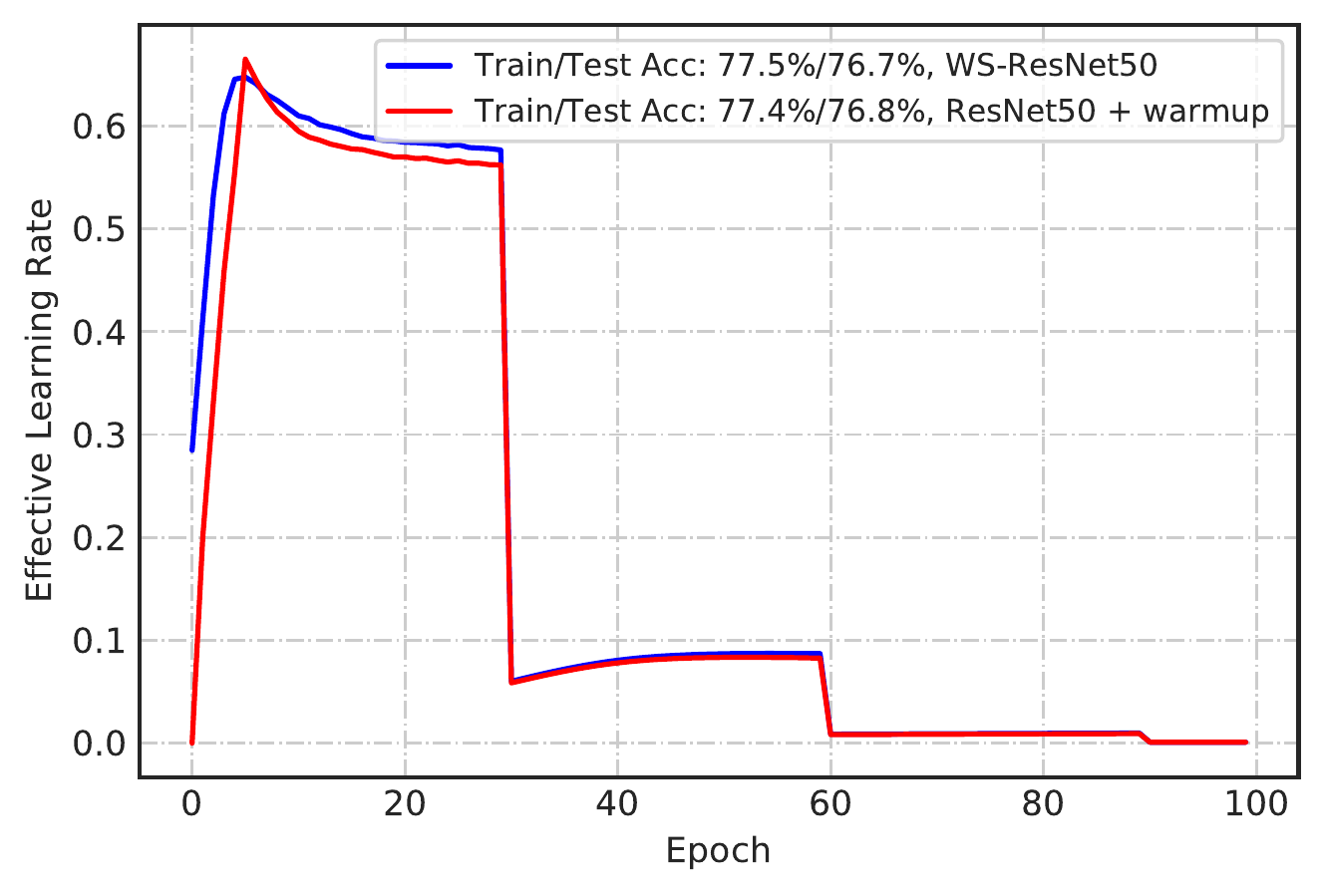}}
	\end{center}	
	\vspace{-16pt}
	\caption{The comparisons of effective learning rate of the first convolutional layer for training WS-ResNet-50 and ResNet-50 with warmup at each epoch during the optimization. The training of WS-ResNet-50 can implicitly simulate the process of warmup to some extent.}
	\label{fig_elrwarmup}
\end{figure}

One more interesting empirical observation is that the control of effective learning rate by weight decay in the early stage is quite similar to a warmup process, and the effectiveness of warmup has been generally confirmed in \cite{he2019bag,you2017large,goyal2017accurate,liu2019variance}. As shown in the Fig. \ref{fig_elrall}, we sample and investigate a set of convolutional layers, and observe that almost all of the layers show an increase in the effective learning rate of several epochs at the beginning of training. The effect of implicit warmup may explain why networks equipped with WS can have slightly improved performance \cite{weightstandardization}. To investigate this deeper, we make additional experiments by explicitly adding warmup process (i.e., linearly increasing the learning rate from 0 to 0.1 during the first 5 epochs) into the training of ResNet-50, and thus partially confirm this conclusion in Table \ref{table_warmup}. Further, the effective learning rate of the first convolutional layer for training WS-ResNet-50 and ResNet-50 with warmup are depicted in Fig. \ref{fig_elrwarmup}. Note that the definition of effective learning rate for ResNet-50 follows \cite{hoffer2018norm}, in order to keep them in a similar magnitude. We suprisingly find that the two curves are very closely matched, which implies that training networks with weight normalization family and weight decay can implicitly have certain benefits of warmup technique. 

\section{Problems via Introducing Weight Decay in Weight Normalization Family}
Despite the certain practical success and benefits of applying traditional weight decay to control the effective learning rate, there are still several serious problems in essence, which are rarely revealed or noticed before our work. In this section, we discuss about these \emph{problems} of introducing weight decay term in the loss objective for weight normalization family in details.
\subsection{No Garantee of Global Minimum}
We first consider WN. If we introduce weight decay term of ${\boldsymbol{W}}$ to the final loss objective, i.e., Eq. (\ref{eq4}), we can prove that for ${\boldsymbol{W}}$, the entire loss function does not theoretically garantee a global minimum. Here we use the proof by contradiction:

If there exists a global minimum ${\boldsymbol{W}}^*$ such that the objective (Eq. (\ref{eq4})) is minimized, then we have the smallest loss $\hat{\mathcal{L}}({\boldsymbol{W}}^*)$ as
\begin{equation}
\hat{\mathcal{L}}({\boldsymbol{W}}^*) = \sum_{i}^N L(f(\boldsymbol{x}_i; \frac{{\boldsymbol{W}}^*}{||{\boldsymbol{W}}^*||}), y_i) + {\frac{1}{2} \lambda ||{\boldsymbol{W}}^*||^2}.
\end{equation} 

Let's take a real number $\alpha (0<\alpha<1)$  and form a new solution ${\boldsymbol{W}}^{\#} = \alpha {\boldsymbol{W}}^*$. Then we have:
\begin{equation}
\begin{aligned}
\hat{\mathcal{L}}&({\boldsymbol{W}}^{\#}) = \sum_{i}^N L(f(\boldsymbol{x}_i; \frac{{\boldsymbol{W}}^{\#}}{||{\boldsymbol{W}}^{\#}||}), y_i) + {\frac{1}{2} \lambda ||{\boldsymbol{W}}^{\#}||^2}\\
&= \sum_{i}^N L(f(\boldsymbol{x}_i; \frac{{\boldsymbol{W}}^{*}}{||{\boldsymbol{W}}^{*}||}), y_i) + {\frac{1}{2} \lambda {\alpha}^2 ||{\boldsymbol{W}}^{*}||^2}\\
&< \sum_{i}^N L(f(\boldsymbol{x}_i; \frac{{\boldsymbol{W}}^*}{||{\boldsymbol{W}}^*||}), y_i) + {\frac{1}{2} \lambda ||{\boldsymbol{W}}^*||^2} = \hat{\mathcal{L}}({\boldsymbol{W}}^*),
\end{aligned}
\end{equation} 
which leads to the contradiction with the assumption that $\hat{\mathcal{L}}({\boldsymbol{W}}^*)$ is smallest. The similar conclusion can be found with WS. 


\subsection{Training Instability}
When given enough training iterations, the objective (Eq. (\ref{eq7})) will continuously push the length of the weight (i.e., $k$) to 0. The effective learning rate is \emph{inversely proportional} to the weight length, which is much easier to cause the floating point overflow and lead to a failed training. 


\begin{table}
	\small
	\centering
	\renewcommand\arraystretch{1.2}
	\newcommand{\tabincell}[2]{\begin{tabular}{@{}#1@{}}#2\end{tabular}}
	\caption{Top-1/5 Accuracy (\%) via single 224$\times$ crop on ImageNet validation set of ResNet-50 with warmup and WS-ResNet-50.}
	\vspace{-10pt}
	\begin{tabular}{l|c}
		\hline
		Type & Top-1/5 Acc (\%)\\ \hline\hline
		{ResNet-50}  &  76.54/93.07 \\ \hline
		ResNet-50 + warmup &  76.81/93.20 \\ \hline
		WS-ResNet-50  & 76.74/93.28  \\ \hline
	\end{tabular}
	
	\label{table_warmup}
\end{table}

\begin{table}
	\small
	\centering
	\renewcommand\arraystretch{1.2}
	\newcommand{\tabincell}[2]{\begin{tabular}{@{}#1@{}}#2\end{tabular}}
	\caption{Top-1 Accuracy (\%) via single 224$\times$ crop on ImageNet validation set of different weight decay $\lambda$ settings for convolution in ResNet-50, WN-convolution in WN-ResNet-50, and WS-convolution in WS-ResNet-50. $\lambda$  of other parts (BN and fc layers) is kept with 1e-4. We demonstrate the results of two widely used optimizers: SGD (with momentum) and Adam. ``--'' denotes a failed training due to the gradient float overflow.}
	\vspace{-10pt}
	\resizebox{0.48\textwidth}{!}{%
	\begin{tabular}{c|c|c|c|c|c|c}
		\hline
		\multicolumn{2}{c|}{Top-1 Acc (\%) w.r.t. $\lambda$} & 1e-2 & 1e-3 & 1e-4 & 1e-5 & 0\\ \hline
		\multirow{2}{*}{ResNet-50}     & SGD  &   47.67   &  74.12    &   76.54   &  74.80  &  72.65 \\ \cline{2-7} 
		& Adam &  19.42  & 35.68 &  52.97   &     63.46   &  72.50 \\ \hline
		\multirow{2}{*}{WN-ResNet-50}  & SGD  &  --    &   --   &  76.44    &  74.65  & 72.86 \\ \cline{2-7} 
		& Adam &   --   &   --   &  --    &  --    & 72.34 \\ \hline 
		\multirow{2}{*}{WS-ResNet-50}  & SGD  &   --   &  --    &   76.74   &  74.70  &  72.92 \\ \cline{2-7} 
		& Adam &  --    &   --   &  --    &   --   & 72.85\\ \hline
	\end{tabular}
	}
	
	
	\label{table_wd}
\end{table}

Specifically, we find that improper selection of $\lambda$ would actually cause the instability in training. When we choose a slightly larger $\lambda$ for an  optimizer, some of the weights in the network will quickly converge to 0, making the effective learning rate close to infinity. Thereby the numerical gradient updates are beyond the representation of float in the computational resource, resulting in a training failure. Table \ref{table_wd} shows the impact of $\lambda$ on network training with two widely used optimizers SGD \cite{sutskever2013importance} with momentum and Adam \cite{kingma2014adam}, where it is much easier to have a failed training for networks with weights normalized. Moreover, the adaptive gradient method (e.g., Adam) even fail to have a successful training unless we discard the weight decay by setting $\lambda = 0$. Since Adam will calculate the square of the gradient during the optimization process, it is more likely to encounter the risk of floating point overflow.

\begin{figure}[t]
	\begin{center}
		\setlength{\fboxrule}{0pt}
		\fbox{\includegraphics[width=0.44\textwidth,height=0.28\textwidth]{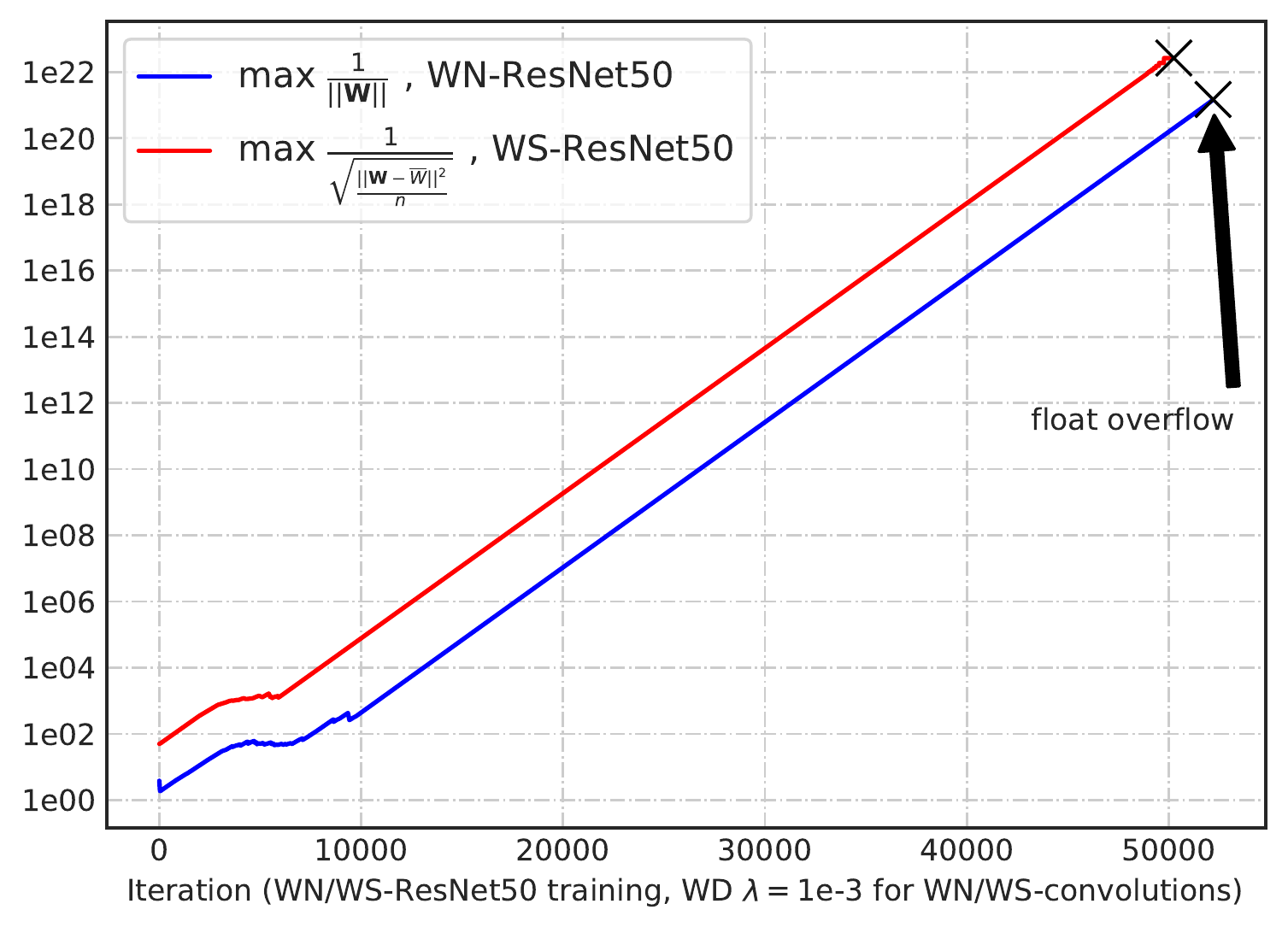}}
	\end{center}	
	\vspace{-16pt}
	\caption{Illustration of gradient float overflow problem over training iterations of WN-ResNet-50 and WS-ResNet-50. We use the maximal reciprocal length (or standard deviation) of weights as statistics.}
	\label{fig_float_overflow}
\end{figure}

To better illustrate the gradient float overflow risks, we demonstrate the maximal $\frac{1}{||\boldsymbol{W}||}$ for WN-ResNet-50 and $\frac{1}{\sqrt{\frac{||{\boldsymbol{W}} - \overline{{{W}}}||^2 }{n}}}$ for WS-ResNet-50 in the case of $\lambda = $ 1e-3 (for all corresponding convolutional layers) during SGD optimization in Fig. \ref{fig_float_overflow}, where the maximal $\frac{1}{||\boldsymbol{W}||}$ (or $\frac{1}{\sqrt{\frac{||{\boldsymbol{W}} - \overline{{{W}}}||^2 }{n}}}$) goes exponentially large and eventually leads to the gradient float overflow after about 50k iterations.

One may argue that the practical implementation of WS \cite{weightstandardization} already considers the risk of float overflow in the original paper by adding a positive constant $\epsilon$ in the denominator of standardization:
\begin{equation}
{\boldsymbol{W}}' = \frac{{\boldsymbol{W}} -\overline{{{W}}}}{\sqrt{\frac{||{\boldsymbol{W}} - \overline{{{W}}}||^2 }{n} } + \epsilon},  \label{eq23}
\end{equation}
here we clarify that only adding $\epsilon$ in the standardization part is definitely not enough for preventing the gradient float overflow problem. To explain, we can derive the gradient of $W'_i$ w.r.t. $W_i$ according to Eq.~(\ref{eq23}):
\begin{equation}
    \frac{1}{\sqrt{\frac{||{\boldsymbol{W}}^{t}\!-\!\overline{{W}^{t}}||^{2}}{n}}\!+\!\epsilon}(1\!-\!\frac{1}{n}\!-\!\frac{\frac{1}{n}({W}_{i}^{t}\!-\!\overline{{W}^{t}})^{2}}{(\sqrt{\frac{||{\boldsymbol{W}}^{t}\!-\!\overline{{W}^{t}}||^{2}}{n}}\!+\!\epsilon)\!\textcolor{blue}{\sqrt{\frac{||{\boldsymbol{W}}^{t}\!-\!\overline{{W}^{t}}||^{2}}{n}}}}),
\end{equation}
where the individual standard deviation term $\textcolor{blue}{\sqrt{\frac{||{\boldsymbol{W}}^{t}\!-\!\overline{{W}^{t}}||^{2}}{n}}}$ still appears in the denominator and the gradient float overflow can still take place consequently. Therefore, it is necessary to propose a different approach to address the problem.

\section{Methods}
This section describes our proposed $\epsilon-$shifted $L_2$ Regularizer in order to address the above problems.

\subsection{$\epsilon-$shifted $L_2$ Regularizer}
As stated in Sec~\ref{sec2.2}, when training weight normalization family with weight decay term, it is suggested to design a mechanism which can successfully prevent the network weights from being extremely small towards 0 in any case of hyper-parameter or optimizer. Given such an insight, we start from investigating the lack of global minimum problem, where Eq.~(\ref{eq7}) is again reviewed:
\begin{align}
& \min_{{\boldsymbol{A}},k}\hat{\mathcal{L}}({\boldsymbol{A}},k)=\sum_{i}^N L(f(\boldsymbol{x}_{i};{\boldsymbol{A}}),y_{i})+\frac{1}{2}\lambda k^{2},\nonumber \\
& s.t.\thinspace\thinspace\thinspace||{\boldsymbol{A}}||=1,k>0. \nonumber
\end{align}

We notice that the central reason for the missing of global minimum is the regular $L_2$ term: $\frac{1}{2}\lambda k^{2} \ (k > 0)$. During optimization, this term will have a large chance to continuously drive $k$ ($k = ||\boldsymbol{W}||$) infinitely close to 0, making the gradient $\frac{\partial \mathcal{L}}{\partial W_i} = \frac{\eta}{||{\boldsymbol{W}}||}(1 - \frac{({ W}_i^{t})^2}{\sum_{j}^n({ W}_j^{t})^2})\frac{\partial \mathcal{L}}{\partial { A}_i}$ to infinity and thus leading to training failures. To avoid such risks, we propose the $\epsilon-$shifted $L_2$ Regularizer, which constrains $k$ from being too small by a positive constant $\epsilon$:

\begin{align}
& \min_{{\boldsymbol{A}},k}\hat{\mathcal{L}}({\boldsymbol{A}},k)=\sum_{i}^N L(f(\boldsymbol{x}_{i};{\boldsymbol{A}}),y_{i})+\frac{1}{2}\lambda (k - \epsilon)^{2},\nonumber \\
& s.t.\thinspace\thinspace\thinspace||{\boldsymbol{A}}||=1,k>0,  \label{eq21}
\end{align}

For the case of WS, given the standard deviation $v > 0$, by modifying Eq.~(\ref{eq10}) we have:
\begin{align}
\min_{{\boldsymbol{B}}, m, v}&\hat{\mathcal{L}}({\boldsymbol{B}}, m, v)  = \sum_{i}^N L(f(\boldsymbol{x}_{i};{\boldsymbol{B}}),y_{i})\!+\!\frac{1}{2} \lambda n(m^2\!+\!(v\!-\!\epsilon)^2),\nonumber \\
& s.t.\thinspace\thinspace\thinspace||{\boldsymbol{B}}||=n,v>0.   \label{eq22}
\end{align}

\subsection{Garantee of Global Minimum}

Thanks to the introduction of $\epsilon-$shifted $L_2$ Regularizer, the modified loss objective (i.e., Eq.~(\ref{eq21}) and Eq.~(\ref{eq22})) now can garantee the existence of global minimum. For WN, suppose $\boldsymbol{A}^*$ is the optimal solution to $\min_{\boldsymbol{A}}{ \sum_{i}^N L(f(\boldsymbol{x}_{i};{\boldsymbol{A}}),y_{i}) }$, then the global minimal solution of $\hat{\mathcal{L}}({\boldsymbol{A}},k)$ is $\boldsymbol{A} = \boldsymbol{A}^*, k = \epsilon$. Therefore, the minimized objective is equal to $\sum_{i}^N L(f(\boldsymbol{x}_{i};{\boldsymbol{A}^*}),y_{i})$, where the additional $\epsilon-$shifted $L_2$ Regularizer $\frac{1}{2}\lambda (k - \epsilon)^{2}$ is utilized during optimization for mainly \emph{two} important purposes: 1) controlling the effective learning rate to help networks converge, and 2) preventing the magnitude of weights from being too small and thus avoiding the gradient float overflow and training failures.

\subsection{Dynamic Decay Mechanism}
In addition, we find that $\epsilon-$shifted $L_2$ Regularizer has the function of dynamically adjusting the decay coefficient according to the current magnitude of training weights during optimization. In the case of WN, the $\epsilon-$shifted $L_2$ Regularizer term is $\frac{1}{2}\lambda (\sqrt{\sum_j^n W_j^2} - \epsilon)^2$, whose gradient formula for $W_i$ is:
\begin{equation}
\lambda(1 - \frac{\epsilon}{\sqrt{\sum_j^n W_j^2}}) W_i = \lambda(1 - \frac{\epsilon}{||\boldsymbol{W}||}) W_i.
\end{equation}

It can also be regarded as an adaptive version of traditional weight decay (i.e., $\lambda W_i$), which uses the dynamic magnitude of training weights $||\boldsymbol{W}||$ to slightly adjust the decay ratio: $\lambda \to \lambda(1 - \frac{\epsilon}{||\boldsymbol{W}||}) $. When the $||\boldsymbol{W}||$ is relatively large, $(1 - \frac{\epsilon}{||\boldsymbol{W}||})$ will also be relatively large, meaning that we will use a larger factor to shrink the larger weights, and it is reasonably intuitive. It probably explains why applying $\epsilon-$shifted $L_2$ Regularizer can have slight improvements in our experiments.

For the case of WS, the gradient formula of $\frac{1}{2} \lambda n(m^2 + (v - \epsilon)^2)$ w.r.t. $W_i$ is:
\begin{equation}
\lambda((1 - \frac{\epsilon}{{\sqrt{\frac{||{\boldsymbol{W}} \!-\! \overline{{{W}}}||^2 }{n}}}})W_i + \frac{\epsilon}{{\sqrt{\frac{||{\boldsymbol{W}} \!-\! \overline{{{W}}}||^2 }{n}}}}\overline{{W}}),
\end{equation}
where the similar analysis can be conducted.

\section{Experiments}
In this section, we conduct extensive experiments on both classification and detection tasks to validate the effectiveness of the proposed $\epsilon-$shifted $L_2$ Regularizer.
\subsection{Experimental Settings}
To validate the effectiveness of the proposed $\epsilon-$shifted $L_2$ Regularizer, we conduct comprehensive experiments on the ImageNet \cite{deng2009imagenet}/CIFAR-100 \cite{krizhevsky2009learning} classification dataset and COCO \cite{lin2014microsoft} detection dataset accordingly. For fair comparisons, all the experiments are run under a unified pytorch \cite{paszke2017automatic} framework, including results of every baseline model. More details can be referred in our public code base: https://github.com/implus/PytorchInsight. We mainly conduct experiments based on the state-of-the-art Weight Normalization (WN) \cite{salimans2016weight} and Weight Standardization (WS) \cite{weightstandardization} from the weight normalization family.

\noindent {\bf ImageNet classification}: The ILSVRC 2012 classification dataset \cite{deng2009imagenet} contains 1.2 million images for training, and 50K for validation, from 1K classes. The training settings for large models are kept similar with \cite{li2019spatial}, except that we set the weight decay ratio $\lambda$ to 0 for all the bias part in networks \cite{he2019bag}, which generally improves about 0.2\% over all the baselines in this paper. We train networks on the training set and report the Top-1 (and Top-5) accuracies on the validation set with single 224 $\times$ 224 central crop. For data augmentation, we follow the standard practice \cite{szegedy2015going} and perform the random-size cropping and random horizontal flipping. All networks are trained with naive softmax cross entropy without label-smoothing regularization \cite{szegedy2016rethinking}. We train all the architectures from scratch by SGD \cite{sutskever2013importance} or Adam \cite{kingma2014adam,loshchilov2018decoupled}. SGD is with weight decay 0.0001 and momentum 0.9 for 100 epochs, starting from learning rate 0.1 and decreasing it by a factor of 10 every 30 epochs. Adam keeps the default settings with learning rate 0.001, $\beta_1 = $ 0.9, $\beta_2 = $ 0.999. The total batch size is set as 256 and 8 GPUs (32 images per GPU) are utilized for training. The default weight initialization strategy is in \cite{he2015delving}, where we use a 'fan\_out' mode specifically. The training settings for small models (i.e., ShuffleNet \cite{zhang2018shufflenet,ma2018shufflenet} and MobileNet \cite{howard2017mobilenets,sandler2018mobilenetv2}) are slightly different according to the references of their original papers \cite{howard2017mobilenets,zhang2018shufflenet}: the default weight decay is 4e-5 with a number of total epochs 300. Warmup \cite{goyal2017accurate}, cosine learning rate decay \cite{loshchilov2016sgdr}, label smoothing \cite{szegedy2016rethinking} and no weight decay on all depthwise convolutional/BN layers \cite{jia2018highly} are as well applied. One should notice that small models are more difficult to train with higher accuracy, so in many papers of small models \cite{howard2017mobilenets,zhang2018shufflenet}, the authors usually take these training tricks as described above. Also importantly, the weight normalization family should not be applied on the \emph{depthwise convolution} (common in those small architectures) in practice since the number of parameters in each normalized group is too small, otherwise we will observe severe performance degradations. In order to make a fair comparison, especially to make the performance of our reimplemented baseline reach the accuracy of the reported ones in the referenced paper, we use these tips in the training of all small models. Note that in our experiments, only those normalized weights are trained with $\epsilon-$shifted $L_2$ Regularizer, others (BN and fc layer weights) are kept with traditional weight decay term (if it exists) since they donot suffer from these problems. 

\begin{table}
	\small
	\centering
	\renewcommand\arraystretch{1.2}
	\newcommand{\tabincell}[2]{\begin{tabular}{@{}#1@{}}#2\end{tabular}}
	\caption{Top-1 Accuracy (\%) via single 224$\times$ crop on ImageNet validation set of different weight decay $\lambda$ settings for WN-convolution in WN-ResNet-50 and WS-convolution in WS-ResNet-50. $\lambda$  of other parts (BN and fc layers) is kept with 1e-4. We demonstrate the results of two widely used optimizers: SGD (with momentum) and Adam. ``+ $\epsilon$'' denotes the use of $\epsilon$-shifted $L_2$ regularizer. }
	\vspace{-10pt}
	\begin{tabular}{c|c|c|c|c|c}
		\hline
		\multicolumn{2}{c|}{Top-1 Acc (\%) w.r.t. $\lambda$} & 1e-2 & 1e-3 & 1e-4 & 1e-5 \\ \hline
		\multirow{2}{*}{WN-ResNet-50 + $\epsilon$}  & SGD  &   71.86   &   75.31   &  76.52    &  74.63  \\ \cline{2-6} 
		& Adam &   64.31   &  64.47     &  65.92    & 68.23   \\ \hline
		\multirow{2}{*}{WS-ResNet-50 + $\epsilon$}  & SGD  &  72.15    &  75.68  &   76.86   & 74.99 \\ \cline{2-6} 
		& Adam &   64.56  &  64.71  &  66.17    &  68.45  \\ \hline
	\end{tabular}
	
	\label{table_train_stable}
\end{table}
\begin{figure}[t]
	\begin{center}
		\setlength{\fboxrule}{0pt}
		\fbox{\includegraphics[width=0.44\textwidth,height=0.28\textwidth]{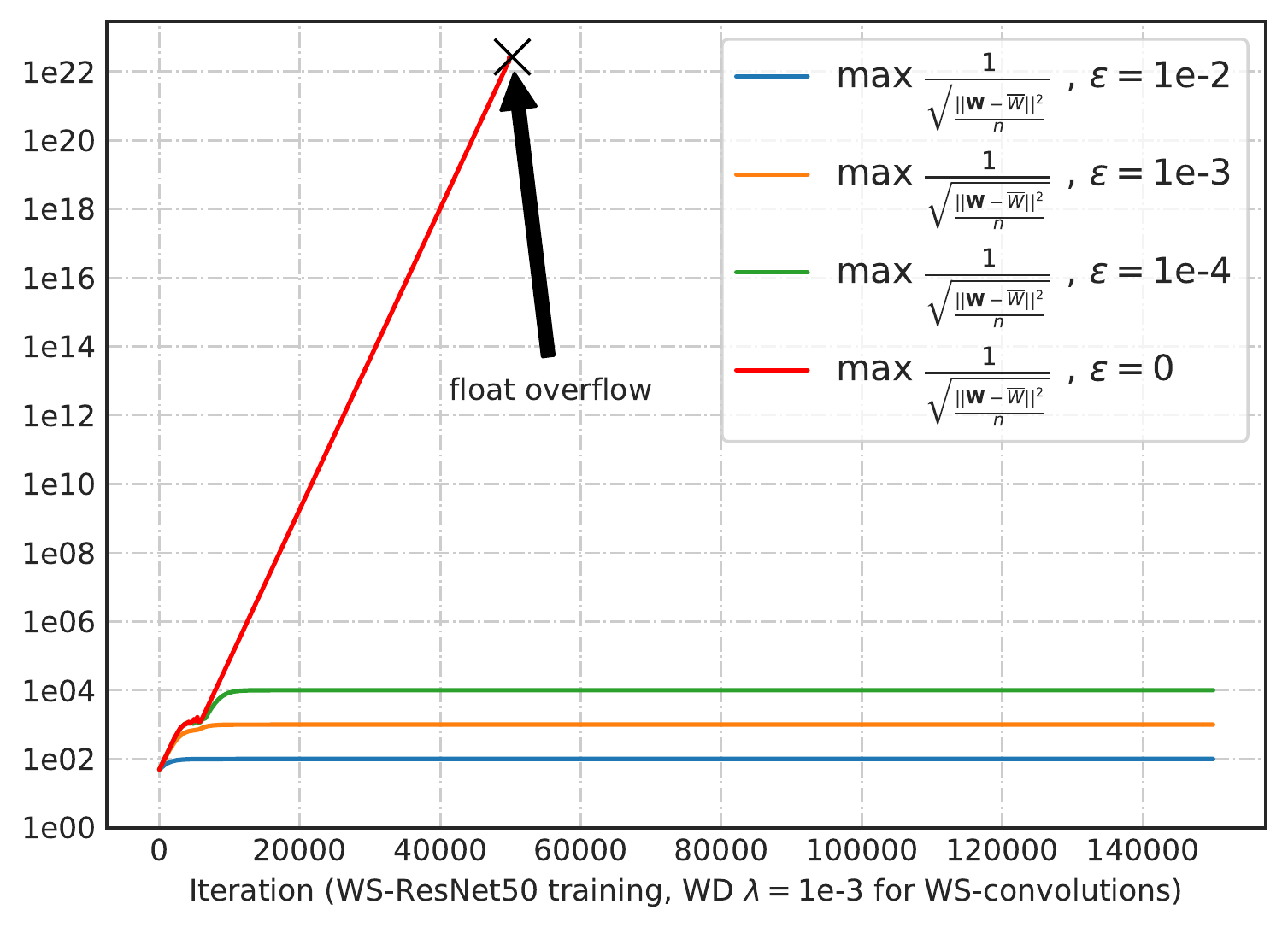}}
	\end{center}	
	\vspace{-16pt}
	\caption{Illustration of training stability over training iterations of WS-ResNet-50 under $\lambda = $ 1e-3. The shifted $\epsilon$ in fact limits the range of gradient float and thus greatly ensures the training stability. ``$\epsilon =$ 0'' denotes the training of WS-ResNet-50 with traditional weight decay.}
	\label{fig_epsilon}
\end{figure}

\noindent {\bf CIFAR-100 classification}: The CIFAR-100 dataset \cite{krizhevsky2009learning} consists of colored natural images with 32 $\times$ 32 pixels, where the images are drawn from 100 classes. The training and test set contain 50K and 10K images, respectively. Apart from the standard data augmentation scheme that is widely used in the dataset \cite{he2016identity,huang2017densely,wang2018mixed}, we also add two recent popular methodologies namely cutout \cite{devries2017improved} and mixup \cite{zhang2017mixup} to further reduce the overfitting risks, where we keep their respective default hyper-parameters in training. For preprocessing, we normalize the data using the channel means and standard deviations. The networks are trained with batch size 64 on one GPU. The training is with weight decay 0.0005 and momentum 0.9 for 300 epochs, starting from learning rate 0.05, which is decreased at 150th and 225th epoch by a factor of 10. 

\noindent {\bf COCO detection}: The COCO 2017 \cite{lin2014microsoft} dataset is comprised of 118k images in train set, and 5k images in validation set. We follow the standard setting \cite{he2017mask} of
evaluating object detection via the standard mean Average-Precision (AP) and mean Average-Recall (AR) \cite{ren2015faster} scores at different box IoUs or object scales, respectively. We use the standard configuration of Cascade
R-CNN \cite{cai2018cascade} with FPN \cite{lin2017feature} and ResNet as the backbone architecture. The input images are resized such that
their shorter side is of 800 pixels. We trained on 8 GPUs with 2 images per GPU. The backbones of all models are pretrained on ImageNet classification, then all layers except for c1 and c2 are jointly finetuned with detection heads. The end-to-end training introduced in \cite{ren2015faster} is adopted for our implementation, which yields better results. We utilize the conventional finetuning setting \cite{ren2015faster} by fixing the learning parameters of BN layers. All models are trained for 20 epochs using Synchronized SGD with a weight decay of 1e-4 and momentum of 0.9. The learning rate is initialized to 0.02, and decays by
a factor of 10 at the 16th and 19th epochs. The choice of
hyper-parameters also follows the latest release of the Cascade
R-CNN benchmark \cite{mmdetection}. For more experiments with other detector frameworks, e.g., Faster R-CNN \cite{ren2015faster} and Mask R-CNN \cite{he2017mask}, we exactly follow the official settings of the baselines described in \cite{mmdetection}.

\begin{table}
	\small
	\centering
	\renewcommand\arraystretch{1.2}
	\newcommand{\tabincell}[2]{\begin{tabular}{@{}#1@{}}#2\end{tabular}}
	\caption{Top-1 Accuracy (\%) via single 224$\times$ crop on ImageNet validation set of different $\epsilon$ settings for WS-convolution in WS-ResNet-50 under $\lambda =$ 1e-4. $\epsilon = 0$ denotes the baseline without the use of $\epsilon$-shifted $L_2$ regularizer. }
	\vspace{-10pt}
	\resizebox{0.48\textwidth}{!}{%
		\begin{tabular}{c|c||c|c|c|c}
			\hline
			Top-1 Acc (\%) w.r.t. $\epsilon$ & 0  & 1e-2 & 1e-3 & 1e-4 & 1e-5 \\ \hline
			WS-ResNet-50 &76.74&76.60&\bf 76.86&\bf 76.84& 76.71 \\ \hline
		\end{tabular}
	}
	\label{table_epsilon}
\end{table}
\begin{table}
	\small
	\centering
	\renewcommand\arraystretch{1.2}
	\newcommand{\tabincell}[2]{\begin{tabular}{@{}#1@{}}#2\end{tabular}}
	\caption{Accuracy via single 224$\times$ crop on ImageNet validation set of different backbones using SGD. $\lambda =$ 1e-4 denotes the results of large models and $\lambda =$ 4e-5 denotes the results of small models. The ``WS'' denotes the conventional weight decay training of WS-equipped networks. ``WS + $\epsilon$'' denotes the introduction of $\epsilon$-shifted $L_2$ regularizer as objectives for training the WS-equipped networks. }
	\vspace{-10pt}
	\begin{tabular}{l|c|c|c}
		\hline
		Top-1 Acc (\%) $\lambda = $ 1e-4 & baseline & WS  & WS + $\epsilon$ \\ \hline\hline
		ResNet-50   \cite{he2016deep}     &    76.54      &    76.74     &  \bf 76.86        \\ \hline
		ResNet-101   \cite{he2016deep}    &    78.17      &    78.07     &  \bf 78.29        \\ \hline
		ResNeXt-50   \cite{xie2017aggregated}    &    77.64      &    77.76    &   \bf 77.88    \\ \hline 
		ResNeXt-101  \cite{xie2017aggregated}    &    78.71      &   78.68     & \bf 78.80 \\ \hline  
		SE-ResNet-101 \cite{hu2018squeeze}   &    78.43      &   78.65     &    \bf 78.75    \\ \hline
		DenseNet-201 \cite{huang2017densely}  & 77.54 & 77.56 & \bf 77.59 \\ \hline\hline
		Top-1 Acc (\%) $\lambda = $ 4e-5 & baseline & WS  & WS + $\epsilon$ \\ \hline\hline
		ShuffleNetV1 1x (g=8) \cite{zhang2018shufflenet}  & 67.62 & 67.84 & \bf 68.09 \\ \hline
		ShuffleNetV2 1x \cite{ma2018shufflenet}  & 69.64 &  69.66 & \bf 69.70 \\ \hline
		MobileNetV1 1x \cite{howard2017mobilenets}    & 73.55 & 73.56 & \bf 73.60\\ \hline
		MobileNetV2 1x \cite{sandler2018mobilenetv2}   & 73.14 & 73.17 & \bf 73.22 \\ \hline
	\end{tabular}
	\label{table_network}
\end{table}

\subsection{Training Stability}
The proposed $\epsilon-$shifted $L_2$ Regularizer ensures the global minimal solution of the magnitude of weights, which is also able to prevent the weights from being too small and thus avoids the gradient float overflow risks. To verify this, we traverse the hyper-parameter $\lambda$ in a large scope and employ $\epsilon-$shifted $L_2$ Regularizer to train networks with weight normalization family (namely, WN and WS), yielding the results in Table \ref{table_train_stable}. In comparison with Table \ref{table_wd}, we find that $\epsilon-$shifted $L_2$ Regularizer not only greatly improves the stability of training, i.e., no matter how the hyper-parameter $\lambda$ changes, the optimization can finally converge to a good solution with no cases of training failures for any type of optimizer, but also slightly boosts the generalization performance in those comparable cases of $\lambda = $ 1e-4 and 1e-5. 

In our experiments, we empirically find that, with $\epsilon-$shifted $L_2$ Regularizer applied, the magnitude of the training weights will actually be controlled in the range of greater than or equal to $\epsilon$ during the entire optimization. To give a better illustration, we depict the maximal $\frac{1}{\sqrt{\frac{||{\boldsymbol{W}} - \overline{{{W}}}||^2 }{n}}}$ for WS-ResNet-50 with $\epsilon-$shifted $L_2$ Regularizer and $\lambda =$ 1e-3 over its training iterations, and vary $\epsilon = $ 1e-2, 1e-3, 1e-4. For a better comparison, we also plot the curve without $\epsilon-$shifted $L_2$ Regularizer (i.e., $\epsilon =$ 0), which is exactly the red curve in Fig. \ref{fig_float_overflow}. As can be seen in Fig. \ref{fig_epsilon}, the shifted $\epsilon$ limits the range of gradient float in fact and thus successfully prevents the training failures.


\subsection{Parameter Sensitivity}
In this subsection, we are interested in the selection of $\epsilon$ as it is the only parameter of the proposed regularizer. To investigate the sensitivity of the hyper-parameter $\epsilon$, we traverse its range in [1e-2, 1e-3, 1e-4, 1e-5] for training WS-ResNet-50 under $\lambda = $ 1e-4, shown in Table \ref{table_epsilon}. It is demonstrated that the choice of $\epsilon$ is robust to the final generalization performance, where more appropriate selections (i.e., $\epsilon = $ 1e-3 and 1e-4) can consistently bring slight improvements of accuracy.

\begin{figure}[t]
	\begin{center}
		\setlength{\fboxrule}{0pt}
		\fbox{\includegraphics[width=0.46\textwidth,height=0.3\textwidth]{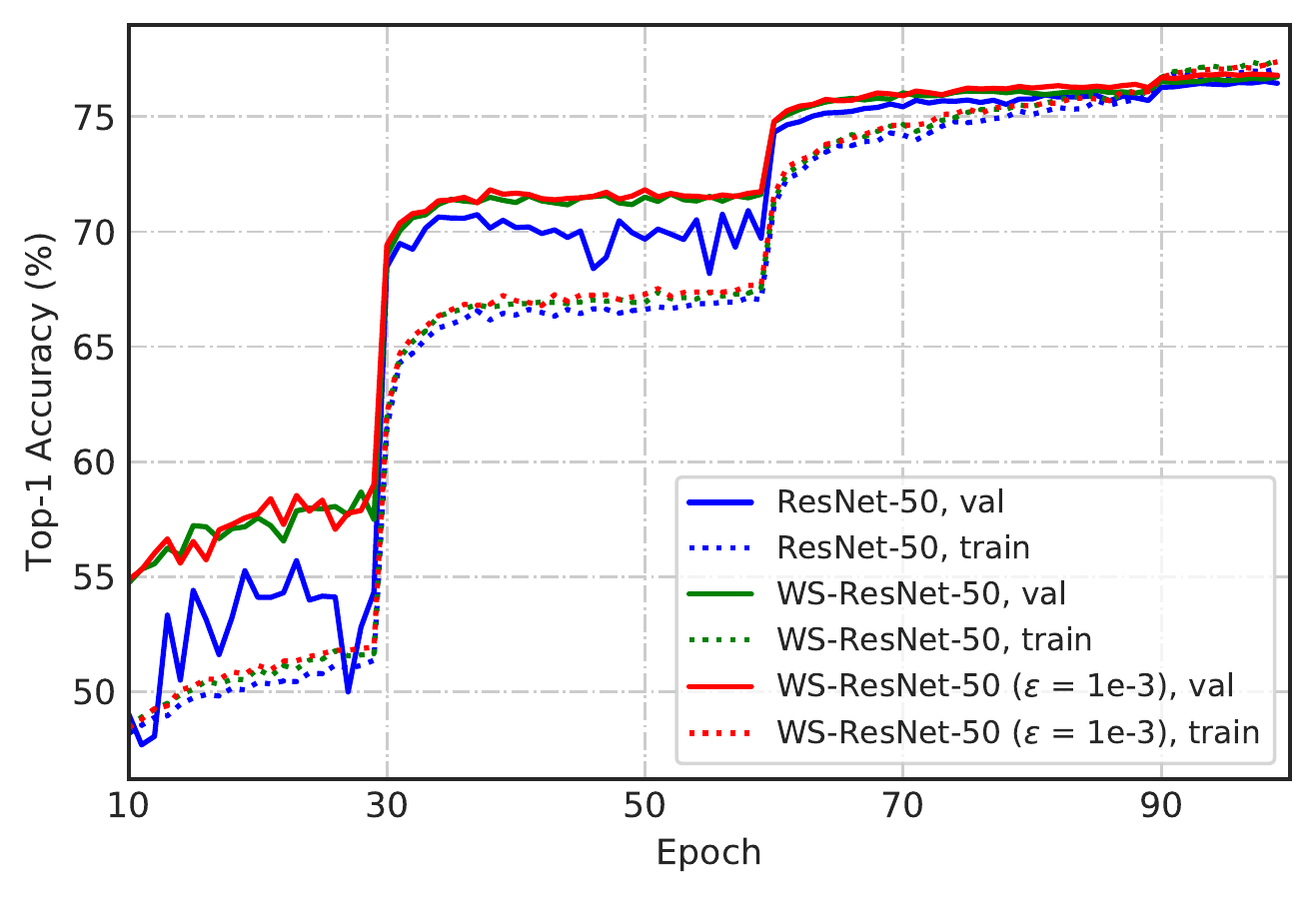}}
	\end{center}	
	\vspace{-16pt}
	\caption{The training and validation curves of (WS-)ResNet-50 on ImageNet dataset. It is observed that the $\epsilon$-shifted $L_2$ regularizer maintains the property of faster convergence.}
	\label{fig_resnet50}
\end{figure}
\subsection{Extension to More Architectures}
Further, we apply $\epsilon$-shifted $L_2$ regularizer to more state-of-the-art network structures \cite{he2016deep,xie2017aggregated,hu2018squeeze,huang2017densely,zhang2018shufflenet,ma2018shufflenet,howard2017mobilenets,sandler2018mobilenetv2} and compare it to the original baseline and traditional WS version using SGD. For the WS-equipped networks, we set hyper-parameters $\lambda =$ 1e-4 and search $\epsilon \in $ \{1e-3, 1e-5, 1e-8\} to report our results. As can be seen from Table \ref{table_network}, while keeping excellent training stability, $\epsilon$-shifted $L_2$ regularizer also achieves very competitive results, both for large and small models. We also empirically demonstrate that the convergence can still be speeded up over the original baseline by $\epsilon$-shifted $L_2$ regularizer in Fig. \ref{fig_resnet50}.

\begin{table}
	\small
	\centering
	\renewcommand\arraystretch{1.2}
	\newcommand{\tabincell}[2]{\begin{tabular}{@{}#1@{}}#2\end{tabular}}
	\caption{Top-1 Accuracy (\%) on CIFAR-100 validation set. For the type, the ``WS'' denotes the conventional weight decay training of WS-equipped networks. ``WS + $\epsilon$'' denotes the introduction of $\epsilon$-shifted $L_2$ regularizer as objectives for training the WS-equipped networks. The numbers in parentheses represent the absolute promotion of baseline.}
	\vspace{-10pt}
	\resizebox{0.48\textwidth}{!}{%
		\begin{tabular}{l|l|l}
			\hline
			Type & Backbone & Top-1 Acc (\%)  \\ \hline
			baseline & ResNeXt29-16x64d \cite{xie2017aggregated} & 83.68 \\ \hline
			WS & WS-ResNeXt29-16x64d \cite{xie2017aggregated} & 83.48 \\ \hline
			WS + $\epsilon$ & WS-ResNeXt29-16x64d \cite{xie2017aggregated} & \bf 84.39 {\scriptsize (+0.71)} \\ \hline\hline
			baseline & SE-ResNeXt29-16x64d \cite{hu2018squeeze}& 84.66 \\ \hline
			WS & WS-SE-ResNeXt29-16x64d \cite{hu2018squeeze}& 84.55 \\ \hline
			WS + $\epsilon$ & WS-SE-ResNeXt29-16x64d \cite{hu2018squeeze}&\bf 84.87 {\scriptsize (+0.21)} \\ \hline
		\end{tabular}
	}
	\label{table_cifar}
\end{table}

\begin{figure}[t]
	\begin{center}
		\setlength{\fboxrule}{0pt}
		\fbox{\includegraphics[width=0.46\textwidth,height=0.3\textwidth]{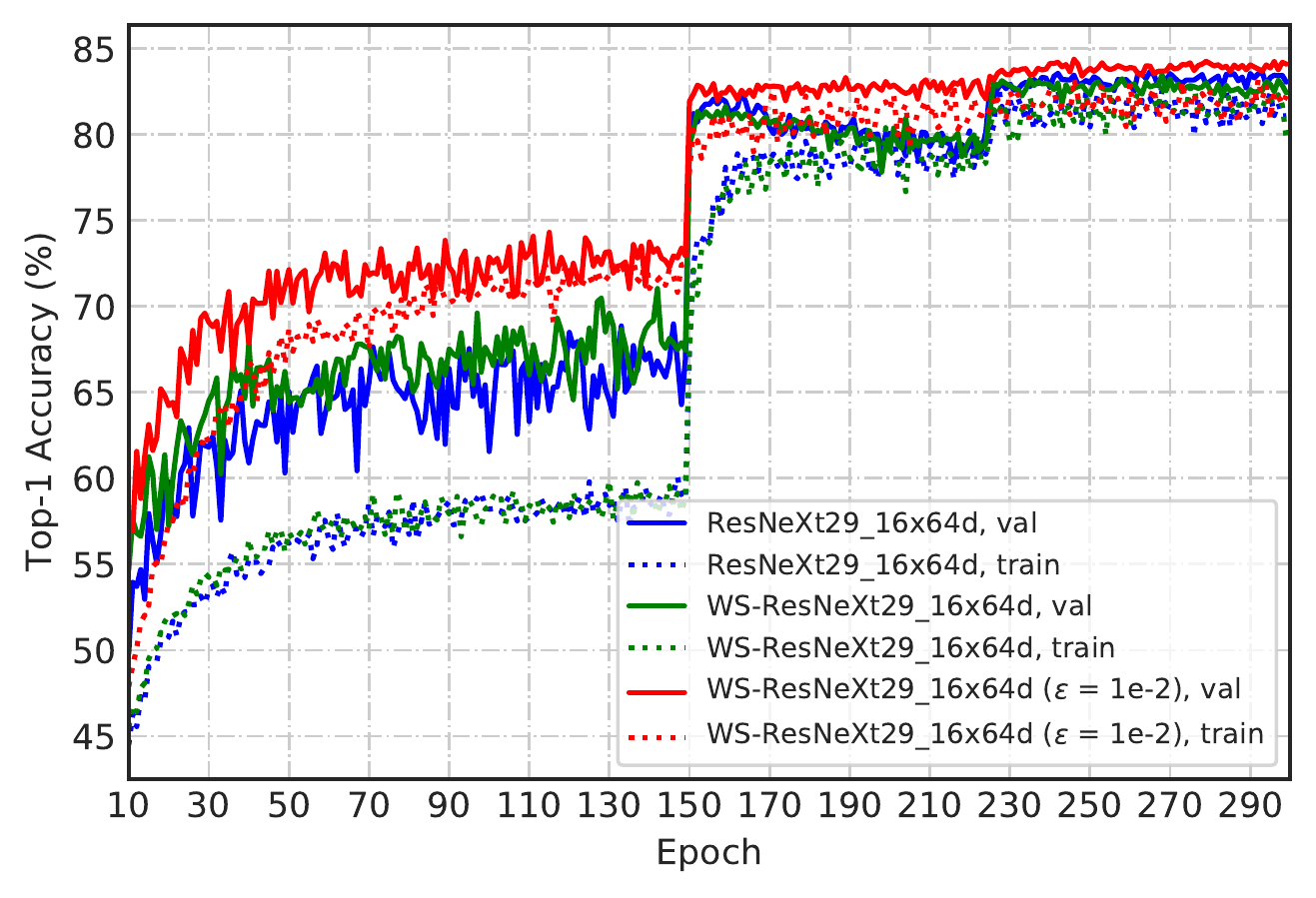}}
	\end{center}	
	\vspace{-16pt}
	\caption{The training and validation curves of different training types of (WS-)ResNeXt29-16x64d on CIFAR-100 dataset. The $\epsilon$-shifted $L_2$ regularizer significantly speeds up the convergence of WS-ResNeXt29-16x64d.}
	\label{fig_cifar}
\end{figure}

\subsection{Extension to Other Datasets }
We further verify whether the effectiveness of $\epsilon$-shifted $L_2$ regularizer can generalise to datasets beyond ImageNet. Here we choose the widely used CIFAR-100, and mainly validate it based on two strong backbones: ResNeXt29-16x64d \cite{xie2017aggregated} and SE-ResNeXt29-16x64d \cite{hu2018squeeze}. In the experiments, we typically find that $\epsilon$ = 1e-2 would bring consistent gains for $\epsilon$-shifted $L_2$ regularizer. From the results in Table~\ref{table_cifar}, we are suprised to discover that in CIFAR-100 for (SE-)ResNeXt29-16x64d, the performance of WS-equipped networks \emph{without} $\epsilon$-shifted $L_2$ regularizer has slightly declined when compared to the baseline. Instead, $\epsilon$-shifted $L_2$ regularizer can still improve the recognition accuracy over the baseline models, which demonstrates its high potentials for practice. Furthermore, the training and validation curves of $\epsilon$-shifted $L_2$ regularizer (i.e., WS + $\epsilon$) can converge significantly faster than baseline and WS, which is depicted in Fig.~\ref{fig_cifar}.

\begin{table*}
	\small
	\centering
	\renewcommand\arraystretch{1.2}
	\newcommand{\tabincell}[2]{\begin{tabular}{@{}#1@{}}#2\end{tabular}}
	\caption{Detection results on COCO 2017 \cite{lin2014microsoft} validation set using Cascade R-CNN \cite{cai2018cascade} and FPN \cite{lin2017feature} with (WS-)ResNet-50 and (WS-)ResNet-101 as backbones. The ``WS'' denotes the conventional weight decay training of WS-equipped networks. ``WS + $\epsilon$'' denotes the introduction of $\epsilon$-shifted $L_2$ regularizer as objectives for training the WS-equipped networks. The numbers in parentheses represent the absolute promotion of baseline.}
	\vspace{-10pt}
	\begin{tabular}{l|l|l|c|c||c|c|c||c|c|c}
		\hline
		Cascade R-CNN \cite{cai2018cascade}  & Backbone & AP & AP$_{.5}$ & AP$_{.75}$ & AP$_s$ & AP$_m$& AP$_l$ & AR$_s$ & AR$_m$& AR$_l$ \\ \hline\hline
		baseline & ResNet-50 & 41.1 & 59.3 & 44.8 & 22.6 & 44.5 & 54.9  & 33.2 & 58.8  & 70.7 \\ \hline
		WS & WS-ResNet-50 &  41.6 & 60.1 & 45.2 & \bf 23.4& 44.7 & \bf 55.6& \bf 34.2 & 58.2 & 71.0\\ \hline
		WS + $\epsilon$ & WS-ResNet-50 & \bf 41.8 {\scriptsize (+ 0.7)} & \bf 60.2 & \bf 45.5 & \bf 23.4 & \bf 45.0 & 55.4 & 33.9 & \bf 58.9 & \bf 71.8\\ \hline\hline
		baseline & ResNet-101        & 42.6 & 60.9 & 46.4 & 23.7 & 46.1 & 56.9 &  34.5 & 59.8 & 72.0 \\ \hline
		WS & WS-ResNet-101              & 43.2 & 61.6 & 47.2 &\bf 24.8 & 46.7 & 57.8 & \bf34.8 & 59.7 &72.2\\ \hline
		WS + $\epsilon$ & WS-ResNet-101 & \bf43.5 {\scriptsize (+ 0.9)} &\bf 61.7 & \bf47.5 & 23.9 &\bf 47.1 & \bf58.4 & 33.4 & \bf60.2 &\bf 72.4\\ \hline
	\end{tabular}
	\label{table_detection}
\end{table*}

\subsection{Extension to Detection Tasks}
We are also interested that whether $\epsilon$-shifted $L_2$ regularizer can still work in some downstream tasks beyond image classification, e.g., object detection. Here we choose one of the most advanced object detector: Cascade R-CNN \cite{cai2018cascade} for evaluation and conduct comprehensive experiments on COCO datasets \cite{lin2014microsoft}. The pretrained models with best performance are utilized for initialization of detector backbones. From Table \ref{table_detection}, it is observed that $\epsilon$-shifted $L_2$ regularizer has the potential to significantly boost the overall performance of the detectors, especially for large backbone ResNet-101 model. Specifically, it purely improves nearly absolute 1\% AP based on the original baseline, and outperforms the baseline in all aspects of other metrics, i.e., AP/AR with different object scales. We also conduct experiments on other state-of-the-art detectors, and observe the consistent improvements in Table \ref{table_more}, which demonstrates its wide usage.

\begin{table}
	\small
	\centering
	\renewcommand\arraystretch{1.2}
	\newcommand{\tabincell}[2]{\begin{tabular}{@{}#1@{}}#2\end{tabular}}
	\caption{Detection results for Faster R-CNN \cite{ren2015faster} and Mask R-CNN \cite{he2017mask} with FPN \cite{lin2017feature} on COCO 2017 \cite{lin2014microsoft} validation set. ``WS + $\epsilon$'' denotes the introduction of $\epsilon$-shifted $L_2$ regularizer as objectives for training the WS-equipped networks.}
	\vspace{-10pt}
	\begin{tabular}{l|l|c|c|c}
		\hline
		Faster R-CNN \cite{ren2015faster}      & Backbone & AP          & AP$_{.5}$ & AP$_{.75}$\\ \hline
		baseline            & ResNet-50        & 37.7 & 59.3 &\bf 41.1\\\hline
		WS + $\epsilon$     & WS-ResNet-50        &\bf 37.9 & \bf59.7 & 40.9 \\ \hline
		baseline            & ResNet-101        & 39.4 & 60.7 & 43.0\\\hline
		WS + $\epsilon$     & WS-ResNet-101        & \bf 39.8 & \bf 60.8 & \bf 43.5 \\ \hline\hline
		Mask R-CNN  \cite{he2017mask}        & Backbone & AP          & AP$_{.5}$ & AP$_{.75}$\\ \hline
		baseline            & ResNet-50        & 38.6 & 60.0 & 41.9\\\hline
		WS + $\epsilon$     & WS-ResNet-50        &\bf 38.9 &\bf 60.1&\bf 42.2 \\ \hline
		baseline            & ResNet-101        & 40.4 & 61.6 & 44.2\\\hline
		WS + $\epsilon$     & WS-ResNet-101        &\bf 41.1  &\bf 62.2 & \bf 45.0 \\ \hline
	\end{tabular}
	\label{table_more}
\end{table}

\subsection{Important Practices for Weight Normalization Family}
In the original papers of weight normalization family \cite{salimans2016weight,weightstandardization}, the authors rarely discuss where to use WN or WS in deep neural networks. Their default mode is to place WN or WS on all conventional convolutional layers, while the BN and fc layers will not participate in WN/WS operations. However, in our practice, it is not always the best option. For depthwise convolutions or group convolutions with very few parameters in each group, using WN or WS can result in a severe degradation of performance both in train and test set. We speculate that when normalizing only a few parameters, since the set of parameters itself has very few degrees of freedom, normalization or standardization will further reduce the degrees of freedom, leading to extremely limited representation ability. One example is the learnable parameter of BN. It is essentially equivalent to a 1$\times$1 depthwise convolution, where each parameter group only contains one variable. If we normalize it, it then becomes a fixed constant (in case of WN), which definitely can not learn the effect of scaling features.

The experiments which we have conducted above mainly avoid these risks. For example, in the small models like ShuffleNetV2, MobileNetV1 and MobileNetV2, we do not apply WS on the depthwise convolutions. And for ShuffleNetV1, it is suggested not to equip the group convolutions with WS. To be specific, we list the results of using or not using WS on the depthwise convolutions in Table \ref{table_ws_where}. It can be observed that whether to use WS on the depthwise convolution can result in a very large performance gap.

\begin{table}
	\small
	\centering
	\renewcommand\arraystretch{1.2}
	\newcommand{\tabincell}[2]{\begin{tabular}{@{}#1@{}}#2\end{tabular}}
	\caption{Top-1/5 accuracy (\%) via single 224$\times$ crop on ImageNet validation set of using or not using WS on the depthwise convolutions. ``w/'' denotes using WS and ``wo/'' denotes not using WS. All other parts are kept the same.}
	\vspace{-10pt}
	\begin{tabular}{l|c|c}
		\hline
		backbone & w/  & wo/ \\ \hline
		WS-ShuffleNetV2 1x \cite{ma2018shufflenet} & 63.79/84.63 & \bf 69.66/88.76 \\ \hline
		WS-MobileNetV2 1x \cite{sandler2018mobilenetv2} & 69.74/89.18 & \bf 73.17/91.05 \\ \hline
	\end{tabular}
	\label{table_ws_where}
\end{table}






\section{Conclusions}
In this paper, we first review the disharmony between weight normalization family and weight decay, i.e., the counter-intuitive under-fitting risk caused by weight decay on the normalized weights. Then, we theoretically answer this question by two evidences: 1) weight decay doesnot change the optimization goal and 2) it ensures the appropriate effective learning rate for better convergence. After that, we expose the detailed problems via introducing fixed weight decay term in the loss objective, including missing of global minimum and training instability. Finally, to solve these potential problems, we propose $\epsilon$-shifted $L_2$ regularizer that shifts the $L_2$ objective by a positive constant $\epsilon$. The shifted $\epsilon$ prevents network weights from being too small, where the gradient float overflow risks can be avoided directly. Comprehensive analyses demonstrate that the proposed  $\epsilon$-shifted $L_2$ regularizer successfully garantees the global minimum and significantly improves the training stability, whilst maintaining superior performance.

\bibliographystyle{plain}
\bibliography{pami_ref}

\end{document}